# Model Interpretation: A Unified Derivative-based Framework for Nonparametric Regression and Supervised Machine Learning

**Xiaoyu Liu[1], Jie Chen, Joel Vaughan, Vijayan Nair, and Agus Sudjianto**
**Corporate Model Risk, Wells Fargo**
**9/7/2018**

## Abstract

Interpreting a nonparametric regression model with many predictors is known to be a challenging problem. There has been renewed interest in this topic due to the extensive use of machine learning algorithms and the difficulty in understanding and explaining their input-output relationships. This paper develops a unified framework using a derivative-based approach for existing tools in the literature, including the partial-dependence plots, marginal plots and accumulated effects plots. It proposes a new interpretation technique called the accumulated total derivative effects plot and demonstrates how its components can be used to develop extensive insights in complex regression models with correlated predictors. The techniques are illustrated through simulation results.

## 1. Introduction

There are many tools in the literature to understand the input-output relationships in nonparametric regression problems, including the "black-box" supervised learning algorithms (Friedman, 2001; Hastie, et al., 2016; Friedman & Popescu, 2008; Apley, 2016; Hu, et al., 2018; Vaughan, et al., 2018). Partial dependence plots (PDP) are the most popular approaches among them. For data with small correlations, PDPs can reliably estimate the relationships between the predictors and fitted response in terms of directions, nonlinearities and interactions. PDPs have been extended to several other diagnostic tools more recently. For example, the Individual Conditional Expectation (ICE) plots were proposed to identify interactions through visualizing the dependence of the predicted response on individual features for each sample point (Goldstein, et al., 2013). The ANOVA decomposition of ICE plots was proposed as a measurement of variable importance which was proved to be consistent with Sobol's global sensitivity indices (Chen, et al., 2018).

However, PDPs and the related tools can be problematic when predictors have moderate to high correlations (Apley, 2016). In such cases, the observed data can be sparse in certain areas of the predictor space, making the nonparametric estimate unreliable outside the support of the observed predictors. Since the PDPs involve extrapolation outside the region of observed data, they can be unreliable as well. Although one can leverage ICE plots to identify the extrapolation of PDPs (Goldstein, et al., 2013; Chen, et al., 2018), the severity of extrapolation cannot be captured by ICE plots. Therefore, it is desirable to have diagnostic tools that can be used more generally, including situations with correlated data.

[1] Email: Xiaoyu.Liu@wellsfargo.com

In this paper, we revisit several other tools that have been proposed in the literature including marginal plots (also known as M plots, bivariate plots or conditional plots) and the ALE plots (Apley, 2016). We provide a unified framework for all these tools, and corresponding theoretical functions, through a derivative-based approach and establish their relationships. Further, we introduce the accumulated total derivative effect (ATDEV) plots which can be decomposed into the existing ALE plots and the newly proposed accumulated cross effects (ACE) plots. This is a new tool that provides useful insights in the regression context when there is dependence among the predictors. Moreover, we demonstrate the usefulness of these components as interpretation tools through simulation studies.

The paper is organized as follows. Section 2 reviews the established interpretation tools including PDPs and marginal plots. Section 3 proposes ATDEV and its decomposition into ALE and ACE. Section 4 revisits PDPs and marginal plots through the derivative-based framework and establishes their relationships with ATDEV plots. Section 5 proposes several tools to visualize the proposed techniques for model interpretation. The use of these tools is demonstrated with simulation examples in Section 6. A set of related derivative-based diagnostics tools is reviewed in Section 7. Section 8 discusses the implementation and computation of ATDEV. A brief conclusion is provided in Section 9.

## 2. Tools for Model Interpretation

Let $Y$ be the response of interest (continuous or binary) and $\boldsymbol{x} = (x_1, \dots, x_p)$ be a $p$-dimensional set of predictors. Further, we have

$$h\big(E(Y|\boldsymbol{x})\big) = f(\boldsymbol{x}) = f(x_1, \dots, x_p), \quad - Eq\ (1)$$

where $f(\cdot)$ represents the regression model and $h(\cdot)$ is a suitable link function. For example, for continuous responses, $h(\cdot)$ can be the identity, logarithm, etc; while for binary responses, $h(\cdot)$ can be logit, probit, etc. The link function $h(\cdot)$ is usually specified or identified by other means. The unknown regression function $f(\cdot)$ is estimated from a sample $\{Y_i,\ \boldsymbol{X}_i = (X_{1i}, \dots, X_{pi}), i = 1, \dots, N\}$.

**Notation:**

We introduce some notations that will be extensively used in the rest of the paper.

a. The sample can be collected according to some pre-specified distribution or, as is more often the case, obtained through some observational mechanism. For both cases, let $g(\boldsymbol{x})$ denote the $p$-dimensional distribution of the predictors. In practice, this can be the empirical distribution of $\{\boldsymbol{X}_i\}$'s. Let $g_j(x_j)$ be the one-dimensional marginal distribution of the $j$-th variable, and $g_{jk}(x_j, x_k)$ be the two-dimensional distribution corresponding to the $(j, k)$-th variables, and so on.
b. $\boldsymbol{x_{-j}} = (x_1, \dots, x_{j-1}, x_{j+1} \dots, x_p)$ is the $(p-1)$-dimensional vector of predictors without the $j$-th variable $x_j$.
c. $\boldsymbol{X_{-j}} = (X_1, \dots, X_{j-1}, X_{j+1} \dots, X_p)$ denote the corresponding random variable.
d. We will also write, with some abuse of notations, $(x_j, \boldsymbol{x_{-j}}) = \boldsymbol{x}$ and $(X_j, \boldsymbol{X_{-j}}) = \boldsymbol{X}$.

There are several one- and higher-dimensional tools that are commonly used to identify and interpret the regression model.

## 2.1. Partial-Dependence Functions and Plots

Perhaps the most common one-dimensional tool is the so-called 1-D partial-dependence (PD) function

$$f_j^{PD}(x_j) = E_{\boldsymbol{X}_{-j}} \{f(x_j, \boldsymbol{X}_{-\boldsymbol{j}})\}. \; - \; Eq\ (2)$$

There are other (some even more reasonable) names in the literature for this function, but since it has become popular in the machine learning (ML) literature, we will refer to it as the 1-D PD function.

If we estimate the response model $f(\boldsymbol{x})$ by $\hat{f}(\boldsymbol{x})$, we can plug it in equation (2) and use the empirical distribution of the data instead of $g(\boldsymbol{x}_{-\boldsymbol{j}})$ to get an estimate of the 1-D PD function. This is referred to as 1-D PD plot (or PDP) in the ML literature. Specifically, the estimate is

$$\hat{f}_j^{PD}(x_j) = \sum_i \hat{f}(x_j, \boldsymbol{x}_{-j,i}) \;/N. \; - Eq\ (3)$$

Here $\boldsymbol{x}_{-j,i}$ denotes $(x_{1,i}, \ldots, x_{j-1,i}, x_{j+1,i}, \ldots, x_{p,i})$. Two- and higher-order PD functions and plots can be obtained similarly.

Let us review some well-known examples:

a) If $f(\boldsymbol{x}) = \beta_0 + \beta_1 x_1 + \cdots \beta_p x_p$ then $f_j^{PD}(x_j) = \beta_j x_j + C_1$ for some constant $C_1$ .
b) The above result holds in general for additive models. That is, if

$f(\boldsymbol{x}) = f_1(x_1) + \cdots + f_p(x_p)$, then $f_j^{PD}(x_j) = f_j(x_j) + C_2$.

c) If $f(\boldsymbol{x}) = \beta_0 + \beta_1 x_1 + \beta_2 x_2 + \beta_{12} x_1 x_2$, then $f_1^{PD}(x_1) = \beta_1 x_1 + \beta_{12}\, \mu_2\, x_1 + C_3$ , where $\mu_2 = E(X_2)$.

If the form of the underlying regression function $f(\boldsymbol{x})$ is unknown, we can use semiparametric or nonparametric methods, including supervised machine learning algorithms, to estimate it. In these cases, there is no analytic form for the estimated PD functions and they have to be displayed as plots. When the correlations among the predictors are small, the observed data will span the $p$-dimensional space well (with the usual caveats about fewer observations in tail regions). If so, the PDPs will provide a reliable estimate of the underlying 1- and higher-dimensional PD functions. However, when (even some of) the correlations are moderate or strong, the support of the observed data can be a much smaller subset of the $p$-dimensional space. If the shape of the underlying regression function $f(\boldsymbol{x})$ varies dramatically, and its behavior inside the data support is different from that outside, the nonparametric estimate $\hat{f}(\boldsymbol{x})$ will be unreliable outside the region of support. The one- and higher-order PDPs involve extrapolation, so they will also be unreliable (see $Eq(3)$). (This problem may not be as serious for parametric models or other well-behaved regression models if they can be well estimated based on just the data within the envelope.) These issues have been recognized in the literature (Apley, 2016; Goldstein, et al., 2013). The diagnostic tools presented in this paper can be used to empirically validate 1-D PDPs in correlated cases. We will discuss this topic in Sections 3 and 4.

## 2.2. Marginal functions and plots

A second class of tools that is also well known is referred to as marginal functions (and their estimates as marginal plots)[2]. Following Friedman (2001), the marginal function can be written as

$$f_j^M(x_j) = E_{\boldsymbol{X}_{-j}|X_j}\{f(X_j, \boldsymbol{X}_{-j})|X_j = x_j\}. - Eq\ (4)$$

Note that the expectation is taken with respect to the conditional distribution of $\boldsymbol{X}_{-j}$ given $X_j$, the $j$-th variable. Thus, the marginal function (and corresponding estimated plots) depend only on the conditional distribution of $\boldsymbol{X}_{-j}|X_j$, a fact that is important when we have highly correlated predictors. On the other hand, recall from $Eq(2)$ that the PD is based on the marginal distribution of $\boldsymbol{X}_{-j}$.

Two-dimensional marginal functions can be defined similarly. They can be estimated by fitting the corresponding models to the data.

When $h(.)$ in $Eq(1)$ is the identity link, we have

$$\begin{aligned} E_{\boldsymbol{X}_{-j}|X_j}\{f(X_j, \boldsymbol{X}_{-j})|X_j = x_j\} &= E_{\boldsymbol{X}_{-j}|X_j}\{E_{Y|\boldsymbol{X}}\{Y|\boldsymbol{X}\}|X_j = x_j\} = \int\int yg(y|\boldsymbol{x})g(\boldsymbol{x}_{-j}|x_j)\,dyd\boldsymbol{x}_{-j} \\ &= \int\int y\frac{g(y,\boldsymbol{x})}{g(\boldsymbol{x})}\frac{g(\boldsymbol{x})}{g(x_j)}dyd\boldsymbol{x}_{-j} = \int\int yg(y,\boldsymbol{x}_{-j}|x_j)\,d\boldsymbol{x}_{-j}dy = \int yg(y|x_j)dy \\ &= E_{Y|X_j}\{Y|X_j = x_j\} \end{aligned}$$

Thus, in this case, the 1-D marginal function can be re-expressed as

$$f_j^M(x_j) = E(Y|x_j), j = 1, \dots, p. -Eq(5)$$

In this important special case, $Eq(5)$ shows that the marginal function corresponds to modeling the response as a function of only the $j$-th variable. The marginal plots have been criticized, and rightly so, because they ignore the effects of the other predictors and hence can be biased. We will return to this point later in the paper.

Since $Eq(5)$ aligns with the concept of simple regression with single predictor, techniques such as LOESS and regression splines for nonparametric regression can be used for the empirical estimation of $f_j^M(x_j)$.

---

[2] As (Apley, 2016) pointed out, the terminology of "marginal plots" is sometimes used to refer to PDPs as the definitions in $Eq(2)$ is with respect to marginal distribution. However, it is more often used to represent the expressions defined in $Eq(4)$, which aims at explaining the response $Y$ or the response surface $f(\boldsymbol{X})$ with single-dimensional input $X_j$. Our paper follows the same definition of marginal plots as in (Apley, 2016).

One of the main contributions in this paper is to further re-express the marginal function as a sum of several components that are useful interpretation tools to understand the dependency among correlated predictors in a regression context. This is the subject of Section 3. In particular, we show that one of the key components in the decomposition is the Accumulated Local Effects (ALE) function and plot proposed in literature (Apley, 2016) for interpreting black box supervised learning models with correlated data. The ALE measures are defined and further discussed in the next section. We will also compare the PD, Marginal, and ALE functions in Section 4 through some well-known examples.

## 3. Accumulated total derivative effects

In this section, we treat the case where all the variables are continuous. The binary case will be disccussed later. Given the $j$-th variable $x_j$, we make the simplifying assumption that for any variable $x_k \in \boldsymbol{x_{-j}}$, its dependence on $x_j$ can be modeled as

$$x_k = m_k(x_j) + e_k,$$

where $e_k$ is random noise with mean 0 and is independent of $x_j$. Then, we can write the first-order total derivative of $f(\boldsymbol{x})$ w.r.t. $x_j$ as

$$f_{j,T}^1(x_j, \boldsymbol{x_{-j}}) = \frac{df(x_j, \boldsymbol{x_{-j}})}{dx_j} = \frac{\partial f(x_j, \boldsymbol{x_{-j}})}{\partial x_j} + \sum_{k \neq j} \frac{\partial f(x_j, \boldsymbol{x_{-j}})}{\partial x_k} \frac{dm_k(x_j)}{dx_j} . -Eq(6)$$

In the rest of the paper, we will denote the first partial derivative of $f(\boldsymbol{x})$ w.r.t. $x_j$ as $f_j^1(x_j, \boldsymbol{x_{-j}})$. Further, we will denote $\frac{dm_k(x_j)}{dx_j}$ as $m_k^1(x_j)$. We can then rewrite $Eq(6)$ as

$$f_{j,T}^1(x_j, \boldsymbol{x_{-j}}) = f_j^1(x_j, \boldsymbol{x_{-j}}) + \sum_{k \neq j} f_k^1(x_k, \boldsymbol{x_{-k}}) m_k^1(x_j) \ . \ - Eq(6')$$

Following the spirit of (Apley, 2016), we integrate $Eq(6')$ to get, what we refer to as, the 1-D Accumulated Total Derivative Effects (ATDEV) of $x_j$:

$$\begin{aligned} f_j^{Tot}(x_j) &= \int_{z_{j,0}}^{x_j} E_{\boldsymbol{X}_{-j}|X_j}\{f_{j,T}^1(X_j, \boldsymbol{X_{-j}}) | X_j = z_j\}\, dz_j \\ &= \int_{z_{j,0}}^{x_j} E_{\boldsymbol{X}_{-j}|X_j}\{f_j^1(X_j, \boldsymbol{X_{-j}}) | X_j = z_j\}\, dz_j \\ &\quad + \sum_{k \neq j} \int_{z_{j,0}}^{x_j} E_{\boldsymbol{X}_{-j}|X_j}\{f_k^1(X_k, \boldsymbol{X_{-k}})\, m_k^1(X_j) | X_j = z_j\}\, dz_j \, , -Eq(7) \end{aligned}$$

where $z_{j,0}$ is some appropriate lower bound. Note from $Eq(7)$ that the right hand side of $f_j^{Tot}(x_j)$ is composed of two parts where the first part is precisely the 1-D Accumulated Local Effect function proposed in Apley (2016):

$$f_j^{ALE}(x_j) = \int_{z_{j,0}}^{x_j} E_{\boldsymbol{X}_{-j}|X_j}\left\{f_j^1(X_j, \boldsymbol{X_{-j}})|X_j = z_j\right\} dz_j. -Eq(8)$$

The second part is a sum of $p-1$ terms, and we refer to each of these as 1-D Accumulated Cross Effects (ACE) which are given by:

$$f_{k,j}^{ACE}(x_j) = \int_{z_{j,0}}^{x_j} E_{\boldsymbol{X}_{-j}|X_j}\left\{f_k^1(X_k, \boldsymbol{X_{-k}})m_k(X_j)\middle|X_j = z_j\right\} dz_j. -Eq(9)$$

Therefore, the 1-D ATDEV function can be written more compactly as

$$f_j^{Tot}(x_j) = f_j^{ALE}(x_j) + \sum_{k \neq j} f_{k,j}^{ACE}(x_j). -Eq(10)$$

This expression indicates that the total 1-D effect of $x_j$ can be decomposed into effects contributed by: i) the partial derivative of $f(\boldsymbol{x})$ with respect to the variable of interest $x_j$, represented by $f_j^{ALE}(x_j)$, and ii) the partial derivatives involving the remaining variables, represented by $f_{k,j}^{ACE}(x_j)$, for $k \neq j$. Note that when $x_j$ is independent of all the other variables, $f_{k,j}^{ACE}(x_j) = 0$, and $f_j^{Tot}(x_j)$ reduces to $f_j^{ALE}(x_j)$, which we will see later is equivalent to $f_j^{PD}(x_j)$ and $f_j^M(x_j)$.

# 4. A unified framework based on the derivatives

## 4.1. Relationships among PDP, marginal and ATDEV plots

The derivative-based approach leads to a unified framework for the PD and Marginal functions discussed in Section 2.

**Proposition 1.**

$$f_j^M(x_j) \equiv f_j^{Tot}(x_j) + C. -Eq(11)$$

**Proof:** See Appendix 1. □

The above result implies that $f_j^M(x_j)$ and $f_j^{Tot}(x_j)$ take different paths to measure the same effect: the 1-D marginal effect of $x_j$ on $f(\boldsymbol{x})$. $f_j^M(x_j)$ directly calculates the conditional averages of the model predictions, whereas $f_j^{Tot}(x_j)$ "detours" by taking derivatives, calculating conditional averages and integrating ("accumulating") the results. The extra effort taken by $f_j^{Tot}(x_j)$ is rewarded by the additional insights and information provided in the ALE and ACE functions. These will be explored in more details soon.

The PD functions can also be unified under the same framework as the marginal functions:

$$f_j^{PD}(x_j) = E_{\boldsymbol{X}_{-j}}\{f(x_j, \boldsymbol{X}_{-j})\} = E_{\boldsymbol{X}_{-j}}\left\{\int_{z_{j,0}}^{x_j} \frac{\partial f(z_j, \boldsymbol{x}_{-j})}{\partial z_j} dz_j + C(\boldsymbol{x}_{-j})\right\}$$

$$= \int_{\boldsymbol{x}_{-j}} \int_{z_{j,0}}^{x_j} \frac{\partial f(z_j, \boldsymbol{x}_{-j})}{\partial z_j} dz_j\, g(\boldsymbol{x}_{-j}) d\boldsymbol{x}_{-j} + C = \int_{z_{j,0}}^{x_j} \int_{\boldsymbol{x}_{-j}} f_j^1(z_j, \boldsymbol{x}_{-j}) g(\boldsymbol{x}_{-j})\, d\boldsymbol{x}_{-j} dz_j + C$$

$$= \int_{z_{j,0}}^{x_j} E_{\boldsymbol{X}_{-j}}\{f_j^1(z_j, \boldsymbol{X}_{-j})\}\, dz_j + C - Eq(12)$$

We see that $f_j^{PD}(x_j)$ is similar to $f_j^{ALE}(x_j)$ as both of them are based on partial derivative $f_j^1(x_j, \boldsymbol{x}_{-j})$. However, $f_j^{PD}(x_j)$ is based on the marignal distribution while $f_j^{ALE}(x_j)$ is based on the conditional distribution. Thus, the latter does not suffer from the extropolation issue in correlated cases that we discussed.

The following results shows that the two are the same in the case of purely additive models.

**Proposition 2.** For a purely additive model given by $f(\boldsymbol{x}) = f_1(x_1) + f_2(x_2) + \cdots + f_p(x_p)$,

$$f_j^{ALE}(x_j) \equiv f_j^{PD}(x_j) + C. -Eq(13)$$

**Proof:**

$$f_j^{PD}(x_j) = E_{\boldsymbol{X}_{-j}}\left\{f_j(x_j) + \sum_{k \neq j} f_k(X_k)\right\} = f_j(x_j) + C_1,$$

$$f_j^{ALE}(x_j) = \int_{z_{j,0}}^{x_j} E\left\{\frac{\partial f(X_j, \boldsymbol{X}_{-j})}{\partial X_j}\middle| X_j = z_j\right\} dz_j = \int_{z_{j,0}}^{x_j} f_j'(z_j) dz_j = f_j(x_j) + C_2,$$

Thus, $f_j^{ALE}(x_j) \equiv f_j^{PD}(x_j) + C$. □

Note that $Eq(13)$ holds for any purely additive function, regardless of the dependence structure among the predictors.

The following result is known (Friedman, 2001). We include it here and provide a proof for completeness.

**Proposition 3.** If the predictors are independent, the PD, Marginal and ALE functions are all equivalent (up to constant differences).

**Proof:**

$$f_j^M(x_j) = E_{\boldsymbol{X}_{-j}|X_j}\{f(X_j, \boldsymbol{X}_{-j}) | X_j = x_j\} = E_{\boldsymbol{X}_{-j}}\{f(X_j, \boldsymbol{X}_{-j})\} = f_j^{PD}(x_j),$$

$$f_j^{ALE}(x_j) = \int_{z_{j,0}}^{x_j} E_{\boldsymbol{X}_{-j}|X_j}\{f^1(X_j, \boldsymbol{X_{-j}})|X_j = z_j\}\, dz_j = \int_{z_{j,0}}^{x_j} E_{\boldsymbol{X}_{-j}}\{f^1(z_j, \boldsymbol{X_{-j}})\}\, dz_j$$
$$= E_{\boldsymbol{X}_{-j}}\left\{\int_{z_{j,0}}^{x_j} f^1(z_j, \boldsymbol{X_{-j}})\, dz_j\right\} = E_{\boldsymbol{X}_{-j}}\{f(x_j, \boldsymbol{X_{-j}}) + C(\boldsymbol{X_{-j}})\} = E_{\boldsymbol{X}_{-j}}\{f(x_j, \boldsymbol{X_{-j}})\} + C$$
$$= f_j^{PD}(x_j) + C,$$

where $C(\boldsymbol{X_{-j}})$ is a function of $\boldsymbol{X_{-j}}$, and $C = E_{\boldsymbol{X}_{-j}}\{C(\boldsymbol{X_{-j}})\}$ is a constant. □

## 4.2. Examples

In this subsection, we illustrate the relationships among 1-D PD, marginal, ALE, ACE, and ATDEV plots through several simple examples with (possibly) correlated predictors. When the predictors are independent, we already know that ACE = 0, and the four measures (PDPs, marginal, ALE and ATDEV) are equivalent.

Let $\boldsymbol{X} = (X_1, X_2)^T$ be bivariate normal with parameters

$$EX_1 = \mu_1, Var(X_1) = \sigma_1^2, \qquad EX_2 = \mu_2, Var(X_2) = \sigma_2^2, \quad Corr(X_1, X_2) = \rho.$$

We can then model $X_1$ as a linear function of $X_2$:

$$X_1 - \mu_1 = \beta_{1|2}(X_2 - \mu_2) + \epsilon_2,$$

where $\beta_{1|2} = \rho\frac{\sigma_1}{\sigma_2}$. Similarly, we have

$$X_2 - \mu_2 = \beta_{2|1}(X_1 - \mu_1) + \epsilon_1$$

with $\beta_{2|1} = \rho\frac{\sigma_2}{\sigma_1}$.

Table 1 summarizes the expressions of the five functions (PD, ALE, ACE, ATDEV and marginal) for the three examples with different functional forms of $f(\boldsymbol{x})$:

**Example 1.** purely additive: $f(x_1, x_2) = x_1 + x_2$
**Example 2.** purely multiplicative: $f(x_1, x_2) = x_1x_2$
**Example 3.** combined: $f(x_1, x_2) = x_1^2 + x_1x_2$

**Table 1. Examples of 1-D PD, ALE, ACE, ATDEV and Marginal functions for different model forms**

| | $\boldsymbol{f(x_1, x_2) = x_1 + x_2}$ | $\boldsymbol{f(x_1, x_2) = x_1x_2}$ | $\boldsymbol{f(x_1, x_2) = x_1^2 + x_1x_2}$ |
|---|---|---|---|
| | | $\boldsymbol{x_1}$ | |
| $f_1^{PD}(x_1)$ | $x_1$ | $\mu_2x_1$ | $x_1^2 + \mu_2x_1$ |
| $f_1^{ALE}(x_1)$ | $x_1$ | $\beta_{2\|1}x_1^2/2 + c_{2\|1}x_1$ | $(1 + \beta_{2\|1}/2)x_1^2 + c_{2\|1}x_1$ |
| $f_2^{ACE}(x_1)$ | $\beta_{2\|1}x_1$ | $\beta_{2\|1}x_1^2/2$ | $\beta_{2\|1}x_1^2/2$ |
| $f_1^{Tot}(x_1) = f_1^M(x_1) + C$ | $(1 + \beta_{2\|1})x_1$ | $\beta_{2\|1}x_1^2 + c_{2\|1}x_1$ | $(1 + \beta_{2\|1})x_1^2 + c_{2\|1}x_1$ |
| | | $\boldsymbol{x_2}$ | |
| $f_2^{PD}(x_2)$ | $x_2$ | $\mu_1x_2$ | $\mu_1x_2$ |
| $f_2^{ALE}(x_2)$ | $x_2$ | $\beta_{1\|2}x_2^2/2 + c_{1\|2}x_2$ | $\beta_{1\|2}x_2^2/2 + c_{1\|2}x_2$ |
| $f_1^{ACE}(x_2)$ | $\beta_{1\|2}x_2$ | $\beta_{1\|2}x_2^2/2$ | $(\beta_{1\|2}/2 + \beta_{1\|2}^2)x_2^2 + 2\beta_{1\|2}c_{1\|2}x_2$ |
| $f_1^{Tot}(x_2) = f_1^M(x_2) + C$ | $(1 + \beta_{1\|2})x_2$ | $\beta_{1\|2}x_2^2 + c_{1\|2}x_2$ | $(\beta_{1\|2} + \beta_{1\|2}^2)x_2^2 + c_{1\|2}(1 + 2\beta_{1\|2})x_2$ |

(In above table, $c_{1|2}$ and $c_{2|1}$ are constants: $c_{k|j} = \mu_k - \beta_{k|j}\mu_j,\ k \neq j$.)

Below are a few comments on these examples:

a. All three examples are consistent with the results in Propositions 1 and 2, and consistent with Proposition 3 when $\rho = 0$.
b. For the purely multiplicative example 2, PD captures only the linear effect of $x_j$, whereas all the other plots indicate quadratic effects for $x_j$. This is because when $x_j$ changes, $x_k$ will also change due to the correlation, and because of the multiplicative effect, this change manifests itself as a quadratic effect. Such quadratic effects are captured by both ALE and marginal functions.
c. When there is both correlation and interaction (multiplicative) (Example 2 or 3), the interaction effects are split into two parts. One part is in ALE (e.g., $\beta_{2|1}x_1^2/2$ in $f_1^{ALE}(x_1)$) as the contribution from the partial derivative w.r.t the variable of interest ($x_j$). The other part is in ACE (e.g., $\beta_{2|1}x_1^2/2$ in $f_2^{ACE}(x_1)$) as the partial derivatives with respect to the other correlated variables $\{x_k\}$'s, or cross effects.

## 5. Visualization Tools

Using the results in Sections 3 and 4, we propose several new visualization tools to aid in understanding the effects and importance of variables in nonparametric regression algorithms, including supervised machine learning. They will be further illustrated with simulation examples in the next section.

### 5.1. ATDEV matrix plot

Given $p$ predictors in the regression model, we create a $p \times p$ matrix plot as follows: The diagonal displays the centered 1-D ALE plots $\left[f_j^{ALE}(x_j) - E\left\{f_j^{ALE}(x_j)\right\}\right]$ for $j = 1, \dots p$. The off-diagonal terms , in row $k$ and column $j$ with $j \neq k$, display the centered 1-D-ACE plots $\left[f_{k,j}^{ACE}(x_j) - E\left\{f_{k,j}^{ACE}(x_j)\right\}\right]$. If the off-diagonal plots are not null, we can make two conclusions: a) the two corresponding variables are dependent; and b) the row variable $x_k$ has a non-zero effect on the fitted response, i.e., the partial derivative $f_k^1(x_k, x_{-k}) \neq 0$. This is because the off-diagonal plots are determined by both the dependence structure $m_k^1(x_j)$ and $f_k^1(x_k, x_{-k})$. If the dependence between $x_j$ and $x_k$ is linear, $m_k^1(x_j)$ is a constant, thus the shape of $f_{k,j}^{ACE}(x_j)$ should "inherit" the shape of the diagonal plot $f_k^{ALE}(x_k)$ for the row variable $x_k$.

### 5.2. Overlay of ATDEV and Marginal plots

The 1-D marginal plot $f_j^M(x_j)$ is usually simply estimated based on the expression in $Eq(4)$. The estimation of 1-D ATDEV plot $f_j^{Tot}(x_j)$ is obtained by adding up the subplots in each column of ATDEV matrix plot. The estimation of $f_j^{Tot}(x_j)$ is much more complicated as it is subject to the approximation errors from derivative calculation and numerical integration, as well as possible biases from the misspecification of $m_k(x_j)$. One or more of these errors can cause the deviation of $f_j^{Tot}(x_j)$ from $f_j^M(x_j)$ empirically. Therefore, we create an overlay of centered 1-D ATDEV plot $[f_j^{Tot}(x_j) - E\{f_j^{Tot}(x_j)\}]$ and centered 1-D marginal plot $[f_j^M(x_j) - E\{f_j^M(x_j)\}]$ for each variable in the model to confirm the estimation accuracy of $f_j^{Tot}(x_j)$ and the related matrix plot in Section 5.1.

## 5.3. Overlay of PDP, ALE and Marginal Plots

For each predictor in the model, we propose overlays of the 1-D PDP, marginal and ALE plots, all centered to remove the level differences from the constant terms. The overlaid plots can give us a sense of overall data correlation, and help to identify PDP extrapolation and its severity in some cases:

- The 1-D marginal curves should be quite different from 1-D PDP and 1-D ALE if there are correlations among the predictors. According to Proposition 3, if one observes the overlap of PDP, marginal and ALE curves for a given variable, it indicates this variable has little correlation with other variables in the model. The larger the difference between marginal and other two curve, the higher the correlation between the variable of interest and other variables.

- If marginal plot does not overlap with the other two curves (i.e., sign of correlation), and there is a good overlap of 1-D PDP and ALE plots for a given variable, according to Proposition 2, this usually indicates this variable's major contribution to the model is additive, and the PDP is not affected too much by the extrapolation issues in spite of the correlation. On the contrary, if PDP and ALE plots are far from each other, it suggests either there are correlated interactions between the variable of interest and other variables, or PDP is contaminated by extrapolations, or both. The extrapolation of PDPs can be further verified by ICE plots (Chen, et al., 2018).

- Once the correlation is identified for a certain predictor through the overlay of PDP, ALE and marginal plots, one can use the correlation matrix to check the sources of the correlation from different predictors, and use ATDEV matrix plot to check the impact of these individual correlations on the fitted response.

## 5.4. Heat Map of ATDEV Components

When the dimension of the predictors $p$ is large, it is difficult to display the $p \times p$ ATDEV matrix plot with $p^2$ curves in total. In such cases, we can summarize the matrix plot into a heat map defined as follows to retain just the primary information.

For subplot $(i, j)$ in the ATDEV plot matrix, we define

$$v_{ij} = \begin{cases} Var\left(f_i^{ACE}(X_j)\right), & i \neq j \\ Var\left(f_j^{ALE}(X_j)\right), & i = j \end{cases}$$

as a measure of the importance of the corresponding effect on the fitted response. Thus each subplot is represented by a single non-negative number, and the whole matrix can be replaced by a heat map, where brighter cells indicate higher levels of importance.

Adding up $v_{ij}$ of each column, we have

$$v_{+j} = \sum_{k=1}^{p} v_{kj},$$

corresponding to the ATDEV importance of each variable in the model. $v_{kj}$ and $v_{+j}$ can be visualized through the heat map or bar charts, respectively. Note that $v_{+j}$ is closely related to, but are not the same as, Sobol indices of 1st order sensitivity defined with $Var\left(f_j^M(X_j)\right)$ (Sobol, 1993).

## 5.5. Heat Maps of ATDEV Components vs Correlation Matrix

The heat map for correlation matrix is well known and easy to understand. It displays the Pearson correlations between all the predictor pairs. Here are the properties of the heat map: (a) it is scaled between -1 and 1; (b) dark colors represent negative correlation and bright colors represent positive correlations; (c) the correlation matrix, and hence the heat map, is symmetric.

The heat map generated from the total derivative components in Section 5.4 is a new tool that provides somewhat different insights into the dependence among predictors. It is more useful in a regression context because it combines the dependence (correlations) among predictors and their influence on the response. This heat map is based on the variance of ALE and ACE functions and has the following properties: (a) the scale is non-negative; (b) dark colors indicate no or little impact and bright colors suggest high impact; (c) the diagonal cells represent the individual marginal contribution of each predictor on the response (i.e., ALE); and (d) the off-diagonal cells represent the magnitude of the cross marginal effect of the column variable on response transferred through the row variable (i.e., ACE). Thus, the diagonal cells of the heat map help to differentiate the effects of the predictors and identify ones with high individual contributions, and an off-diagonal cell which is not completely dark indicate the corresponding column variable has some impact on response through the intermediate row variable. The brighter the cell is, the higher the transferred impact.

Unlike the heat map for correlation matrix, the new heat map is asymmetric, because in theory the off-diagonal cells depend not only on correlation but also on the magnitude of partial derivatives of the response surface with respect to the intermediate row variables, which can be different between the two variables of a pair. See Table 2, Figure 1.a and Figure 1.b for the comparison summary between the heat map for the correlation matrix and the heat map of ATDEV components.

In the heat map of ATDEV components, the "total brightness" in each column indicates the overall "marginal" impact of each predictor on response. And the distribution of the bright cells in each column indicates the different sources of such overall impact. In variable selection, ideally one would want to keep the variables with the large overall marginal effect as well as large individual contribution (i.e., the bright diagonals). For example, if two variables have similar overall marginal brightness, the one with bright diagonal cell is more preferable in the model as its overall contribution comes mainly from itself and is less affected when its correlated variables are removed from the model. By comparison, the other candidate variable relies more on the intermediate variables to pass its impact on to response. The distribution of the bright cells for such candidate column variable is usually highly sensitive to samples and algorithms. This is similar to the "shared significance" phenomenon in linear regression for confounded variables. Even if such candidate is removed from the model, it is likely that its contribution to response will be taken over by its correlated variables still in the model. Therefore such candidate is less "important" than its counterpart with bright diagonal cells.

The figure below shows examples of the correlation matrix heat map (left) and ATDEV decomposition heat map (right) based on a home lending modeling data with 24 predictors. In this example, var553, var550 and var556 are good demonstrations of the above concept. They have similar "sum of brightness" scores in their corresponding columns (see Figure 1.b), whereas var553 and var556 are much brighter than var550 in diagonal cells, indicating stronger individual 1-D effects. We also noticed that the brightness of several non-dark cells in the column of var550 varies across different runs of NN algorithms, indicating the instability of the shared significance due to variable correlations. On the other hand, var553 and var556's column patterns are more stable than that of var550, where the diagonal cell always dominates the column brightness, indicating stable individual contribution less affected by correlations.

**Table 2. Comparison between correlation matrix heat map and ATDEV components heat map**

| | **Correlation Matrix** | **Total Derivative Decomposition** |
|---|---|---|
| **Meaning** | Pearson correlations among **X**'s | **X**'s impact on Y, or 1-D global sensitivity |
| **Type** | Unsupervised analysis | Supervised analysis |
| **Use** | Correlation identification | Identify effective correlation, help with variable selection |
| **Scale** | [-1,1] | [0,+Inf) |
| **Diagonal cells** | All have scale = 1 | Different scales indicate different levels of individual contribution (ALE) |
| **Off-diagonal cells** | Symmetric at diagonal | Asymmetric, indicate different levels of transferred cross effects through correlations (ACE) |

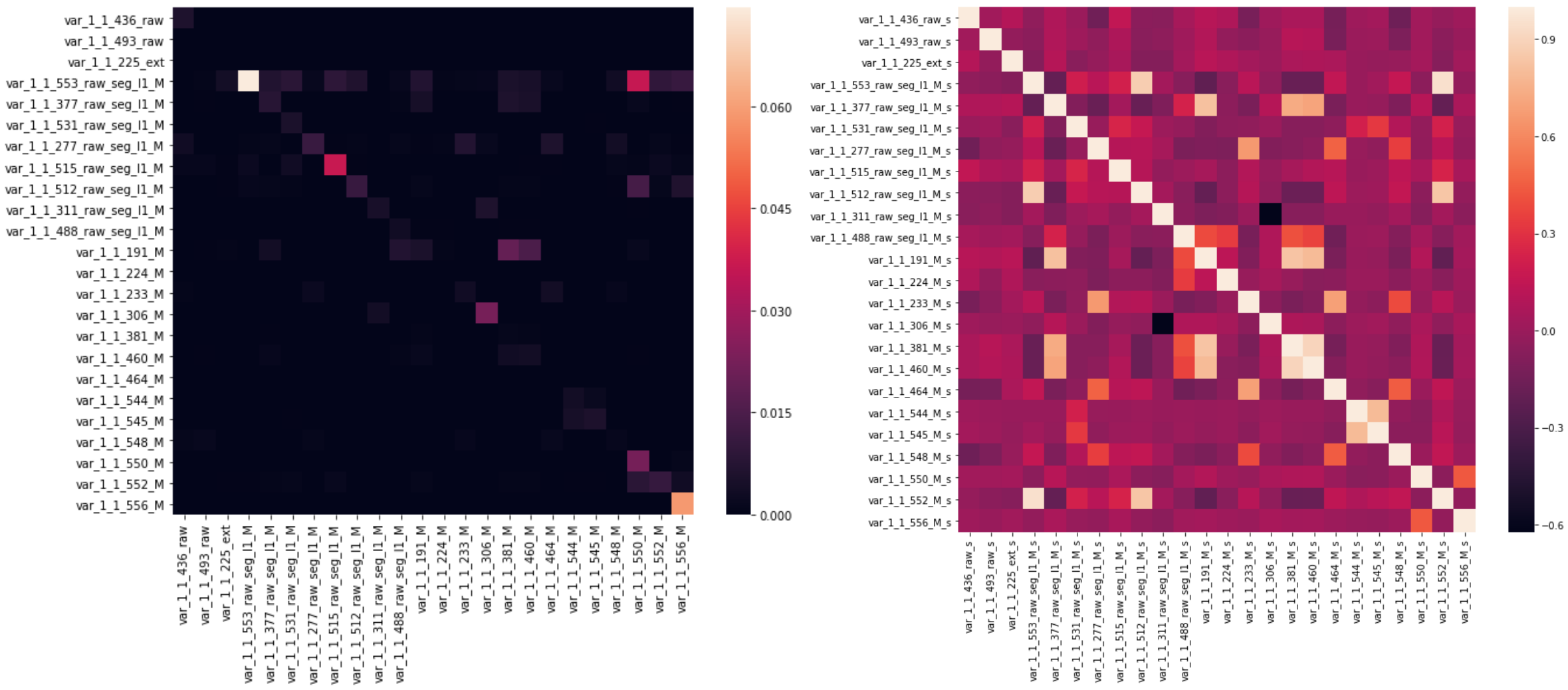

Figure 1.a. Illustrative example of ATDEV components heat map (left) and correlation heat map (right)

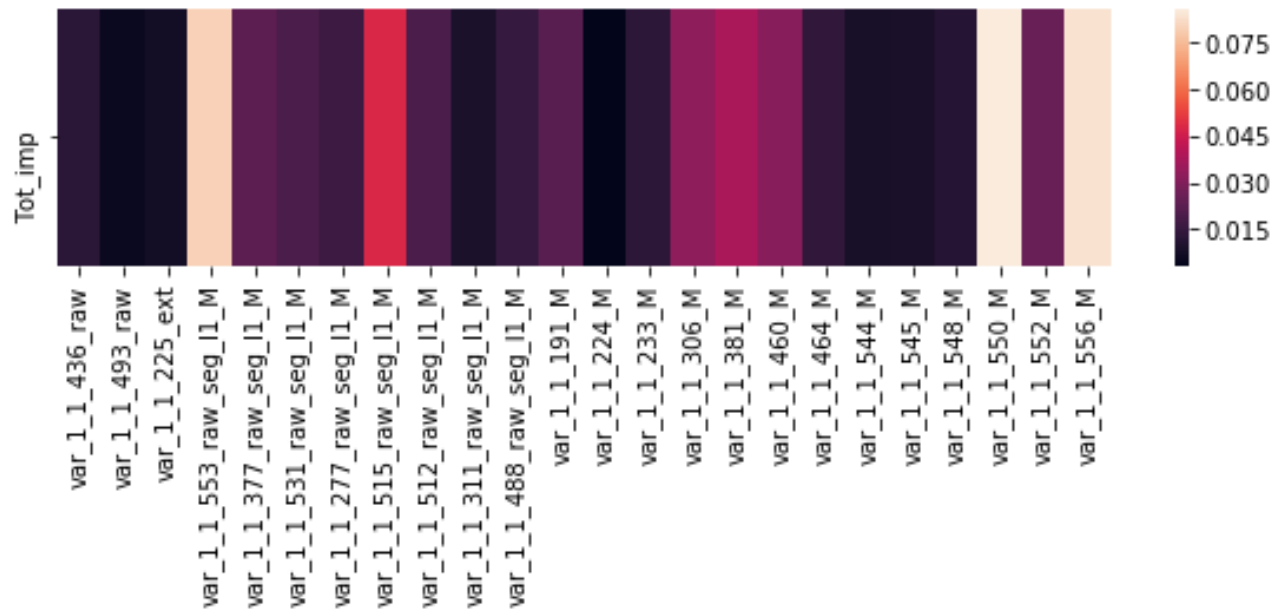

Figure 1.b. Total 1-D Variable Importance ($v_{+j}$) heat map from the above ATDEV components heat map

# 6. Simulation Study

This section illustrates the visualization tools discussed in Section 5 using results from simulation studies. For each of the simulation cases, we generated n=100,000 training samples following a specific functional form, and fitted a Feedforward Neural Network (NN) with 1 hidden layer and linear output activation function. To control overfitting, we generated n=2,000 validation samples, applied early stopping based on the validation MSE, and ensured a close match between the model fit $R^2$ with the theoretical $R^2 = 1 - var(\epsilon)/var(Y)$ on the validation data.

## 6.1. Independent case

We generated $X_1, X_2, X_3, X_4, X_5 \sim Unif(-1,1), \epsilon \sim N(0,0.1^2)$, all independent.

$$y = x_1 + x_2^2 + x_3^3 + 0.8x_2x_4 + \epsilon$$

The fitted NN has validation $R^2 = 0.984$, close to the theoretical $R^2 = 0.986$ for the validation data. The figures below show the three sets of diagnostic plots: (a) ATDEV plot matrix, (b) ATDEV and marginal plot overlay, and (c) PDPs, marginal and ALE plots overlay.

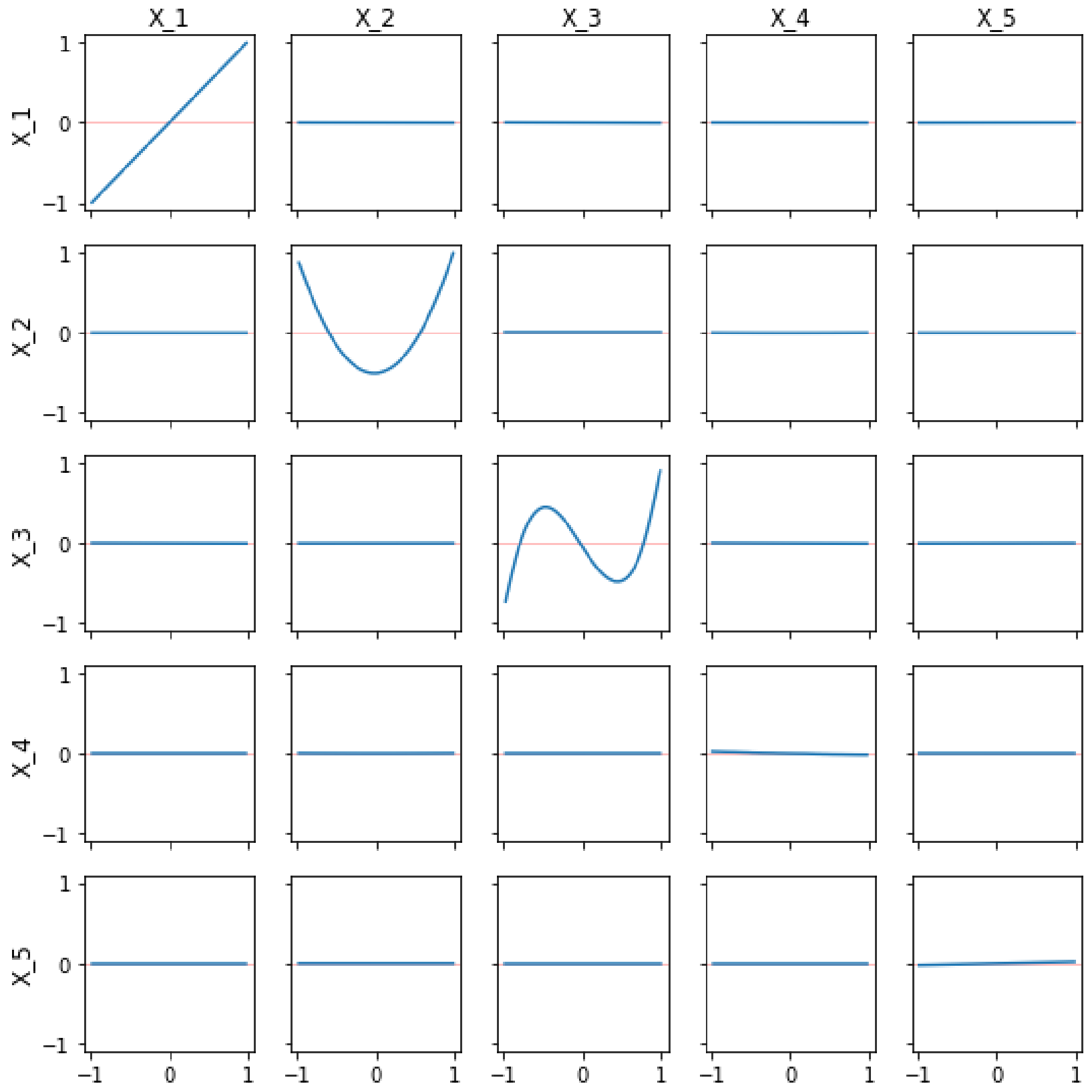

**Figure 2.a. ATDEV matrix plot**

In the independent case, the 1-D ALE plots (i.e., the diagonals of the ATDEV plot matrix) truly reflect the main effects of $x_1, x_2, x_3$, respectively. Since there is no correlation between any variables in the model, the ACE plots (i.e., the off diagonals of the ATDEV plot matrix) are all zero in Figure 2.a, and PDPs, marginal and ALE plots have a good overlap for all of $x_1, x_2$ and $x_3$ in Figure 2.c. The blue curves in ATDEV plots in Figure 2.b are the sum of each column in Figure 2.a and they overlap well with the marginal plots in green(ish) curves, suggesting that the calculations of ALE and ACE plots in Figure 2.a are reliable.

Note that the interaction effects of $x_2$ and $x_4$ are not captured in any of these 1-D plots which only represent the 1-D marginal effects of the variables. One can refer to the 2-D plots to capture such pure 2-way interactions not confounded with any correlations (Apley, 2016). One can also use LE plots introduced in Section 7 to detect interactions for the independence case.

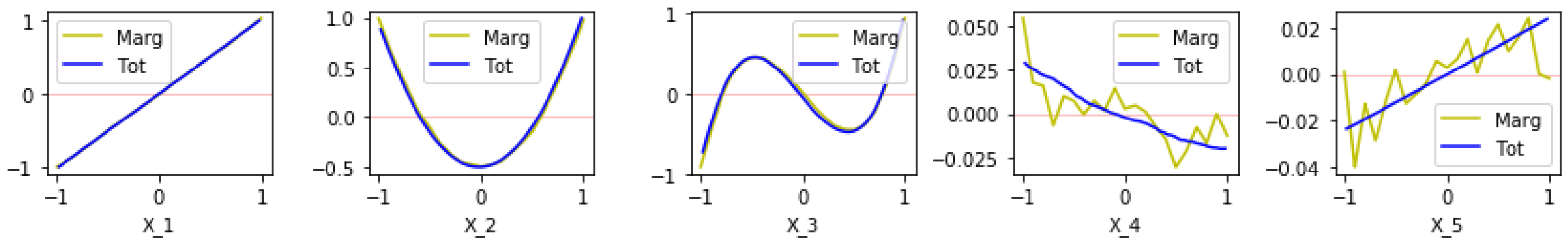

**Figure 2.b. ATDEV and marginal plot overlay**

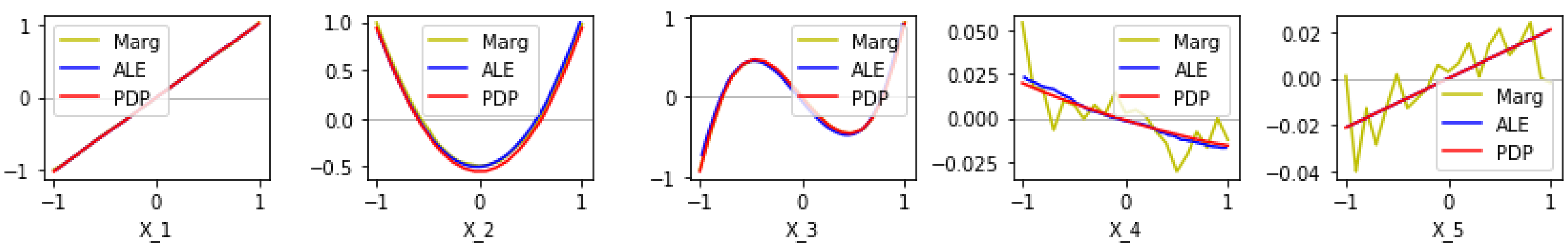

Figure 2.c. PDPs, marginal and ALE plots overlay

## 6.2. Correlated continuous cases

In this subsection we show three simulation cases, all with continuous correlated variables, one with simply purely additive functional form and two with additive terms plus interaction.

**Case 1. Simple additive**

We generate $x_1 \sim Unif(-1,1)$, $\epsilon \sim N(0,0.1^2)$, which are all independent; we also generate $x_2$ and $x_3$ through the following:

$$x_2 = 0.8x_1 + \epsilon, \quad x_3 = -x_1 + \epsilon$$

and

$$y = x_1^2 + \mathrm{x}_2 + \epsilon.$$

The sample correlations of the data are shown in Table 3. There are strong correlations among $x_1$, $x_2$ and $x_3$ .

Table 3. Correlation table

| | x1 | x2 | x3 |
|---|---|---|---|
| **x1** | 1.000000 | -0.962662 | -0.985348 |
| **x2** | -0.962662 | 1.000000 | 0.977116 |
| **x3** | -0.985348 | 0.977116 | 1.000000 |

The fitted NN has $R^2 = 0.969$ on the validation data, close to the theoretical $R^2 = 0.970$.

In Figure 3.a, the quadratic main effect term of $x_1$ and the linear main effect term of $x_2$ are correctly captured by ALE plots. $x_3$ does not have any direct main effect term in the model, thus its ALE is constant at 0. The off-diagonal ACE plots represent the indirect 1-D marginal effects through correlations. For example, subplot (2,1) shows $x_1$'s effect on the response surface through its correlation with $x_2$. Since $x_2$ itself has linear individual effect (see subplot (2,2)), and $m_2^1(x_1)$ is a constant, the shape $f_2^{ACE}(x_1)$ in this off-diagonal cell resembles that of the diagonal $f_2^{ALE}(x_2)$ in the same row. The same is true for $f_1^{ACE}(x_2)$.

Note that although $x_3$ does not have any direct 1-D main effect term in the model, since it has a negative correlation with both $x_1$ and $x_2$, and both $x_1$ and $x_2$ are in the model, $x_3$ has strong indirect effects transferred through $x_1$ and $x_2$. It is quite obvious that the negative relationship between $x_2$ and $x_3$ leads to the opposite directions of the straight lines for $f_2^{ALE}(x_2)$ and $f_2^{ACE}(x_3)$. Let $E(X_1|X_3) = \beta_{3|1}X_3$ where $\beta_{3|1}$ is -1 in our setting , and we can tell the reason why the quadratic curves of $f_1^{ACE}(x_3)$ and $f_1^{ALE}(x_1)$ open to the same direction from the definition of ACE,

$$f_1^{ACE}(x_3) = \int_{z_{3,0}}^{x_3} E(f_j^1(X_1, \boldsymbol{X^{-1}})\,\beta_{3|1}|X_3 = z_3)dz_3 = \int_{z_{3,0}}^{x_3} E(2\beta_{3|1}X_1\,|X_3 = z_3)dz_3 = \int_{z_{3,0}}^{x_3} 2\beta_{3|1}^2 z_3\,dz_3$$
$$= \beta_{3|1}^2 x_3^2 + C.$$

Figure 3.b shows the overall effect of each variable on the response surface. Note that the 3 variables have similar shapes of total effects due to their high correlations and ATDEV plots overlap well with the marginal plots. The only exception for good overlapping is the right tail part for $x_2$ which is caused by the numerical approximation inaccuracies of both the marginal plot and ATDEV plot due to sparse data samples.

Due to the correlations, marginal plots no longer overlap with PDPs and ALE plots in Figure 3.c. However, since the functional form is purely additive, there is a close match between PDPs and ALE plots according to proposition 2. Although this example does not show this, it is worth mentioning that if one observes big gaps between PDPs and ALE plots for purely additive models, it suggests that the PDPs have extrapolation issues.

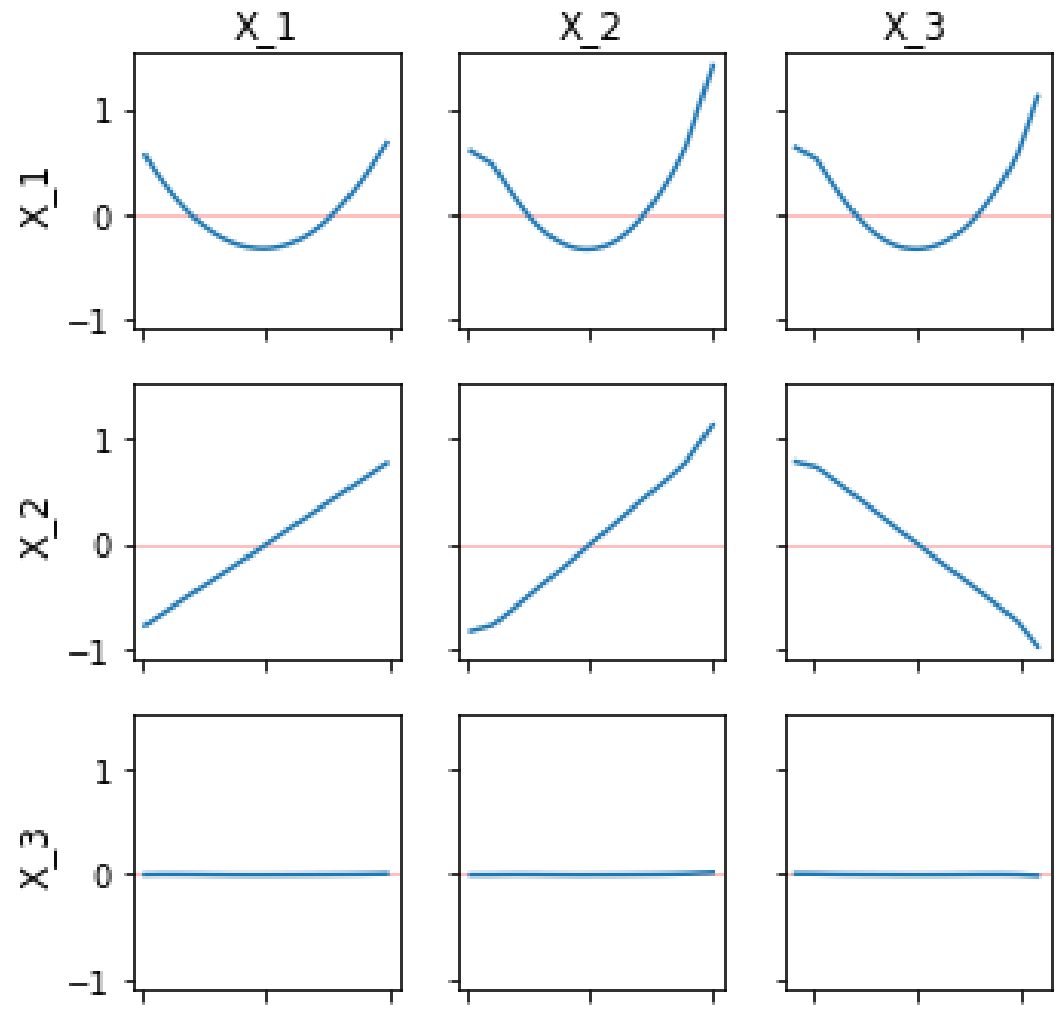

**Figure 3.a. ATDEV matrix plot**

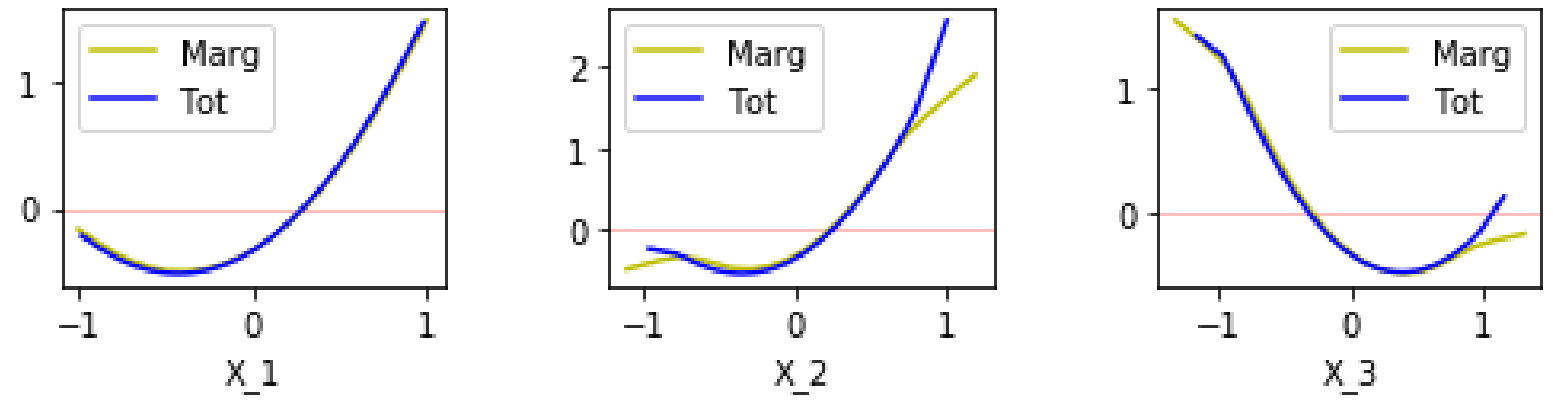

**Figure 3.b. ATDEV and marginal plot overlay**

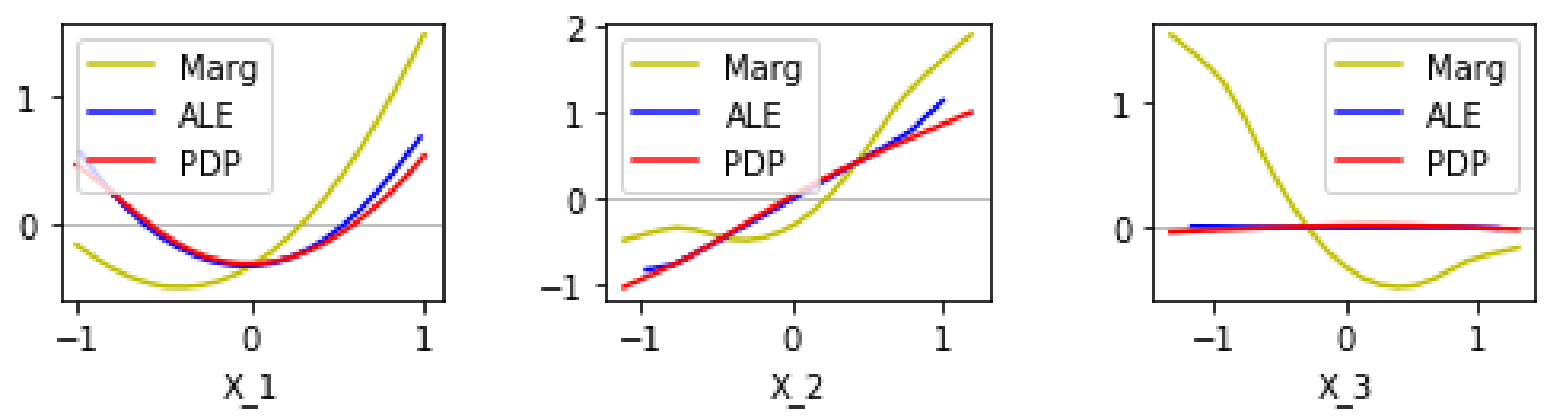

Figure 3.c. PDPs, marginal and ALE plots overlay

**Case 2. Simple additive plus interaction**

We generate $x_1, x_3 \sim Unif(-1,1)$, $\epsilon \sim N(0, 0.1^2)$, which are all independent; we also generate $x_2$ through the following:

$$x_2 = -x_1 + \epsilon$$

and

$$y = x_1 + x_2 + x_1 x_2 + \epsilon$$

In this case, $x_1$ and $x_2$ have a strong negative correlation of -0.985. NN has $R^2 = 0.911$ with the validation dataset, close to the theoretical $R^2 = 0.912$.

In Section 4.2, we explain how the two-way interaction $x_i x_j$ generates the quadratic ALE plots through correlations. Such effect is well captured in Figure 4.a, where the diagonal plots (ALE) show the quadratic effects of $x_1$ and $x_2$ from the correlated interaction $x_1 x_2$. By combining Example 1 and 2 in Section 4.2, and considering $\mu_1 = \mu_2 = 0$, the diagnostics plots have the following analytic forms:

$$f_1^{PD}(x_1) = x_1, \qquad f_2^{PD}(x_2) = x_2$$

$$f_1^{ALE}(x_1) = \frac{\beta_{2|1}}{2} x_1^2 + x_1, \quad f_2^{ALE}(x_2) = \frac{\beta_{1|2}}{2} x_2^2 + x_2$$

$$f_2^{ACE}(x_1) = \frac{\beta_{2|1}}{2} x_1^2 + \beta_{2|1} x_1, \quad f_1^{ACE}(x_2) = \frac{\beta_{1|2}}{2} x_2^2 + \beta_{2|1} x_2$$

$$f_1^{Tot}(x_1) = f_1^M(x_1) = \beta_{2|1} x_1^2 + \left(1 + \beta_{2|1}\right) x_1, \quad f_2^{Tot}(x_2) = f_2^M(x_2) = \beta_{1|2} x_2^2 + \left(1 + \beta_{1|2}\right) x_2$$

The OLS estimates $\hat{\beta}_{2|1} = \hat{\beta}_{1|2} = -0.98$ are used in generating the following plots. All the estimated ALE, ACE, ATDEV and marginal plots match with their theoretical forms. We notice that the estimated PDPs in Figure 4.c have a little curvature, whereas their analytic forms are linear indicating that the inconsistency is caused by the extrapolation issue of the correlated data.

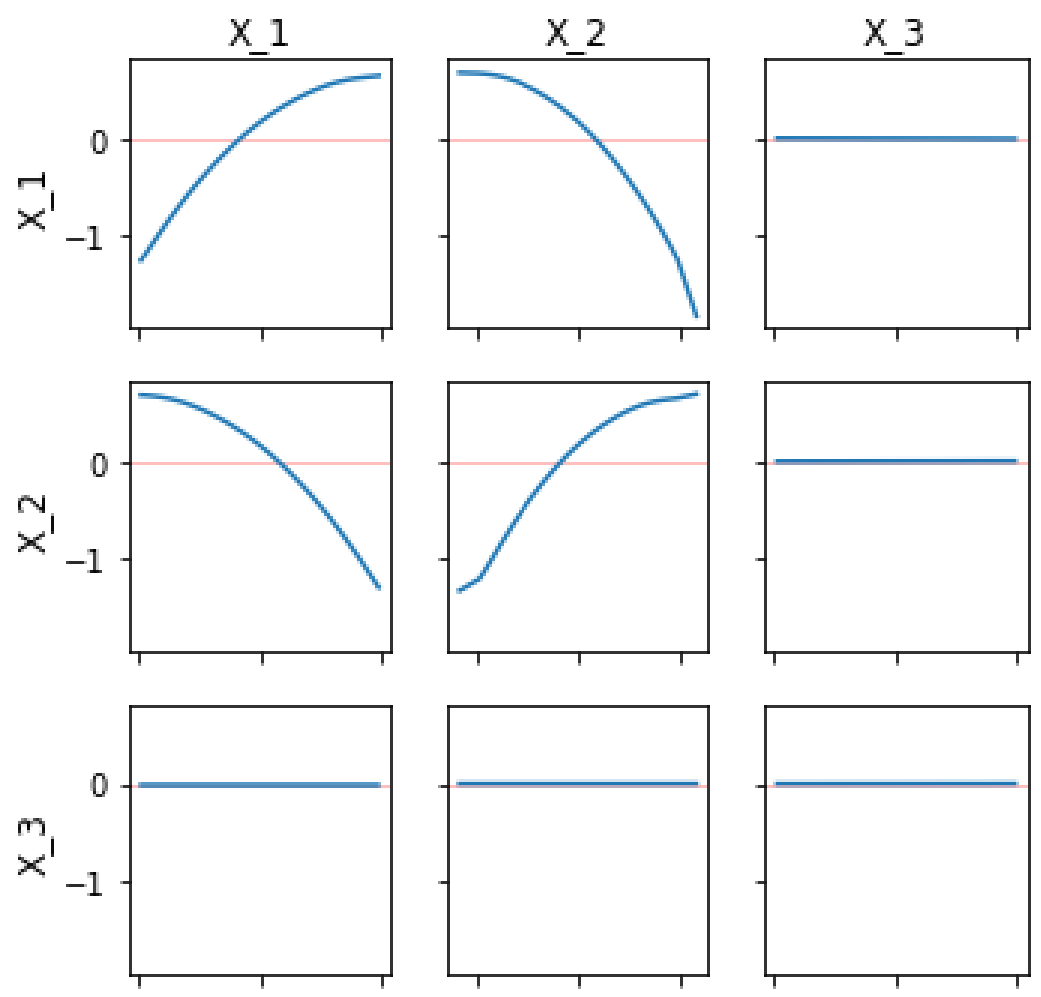

**Figure 4.a. ATDEV matrix plot**

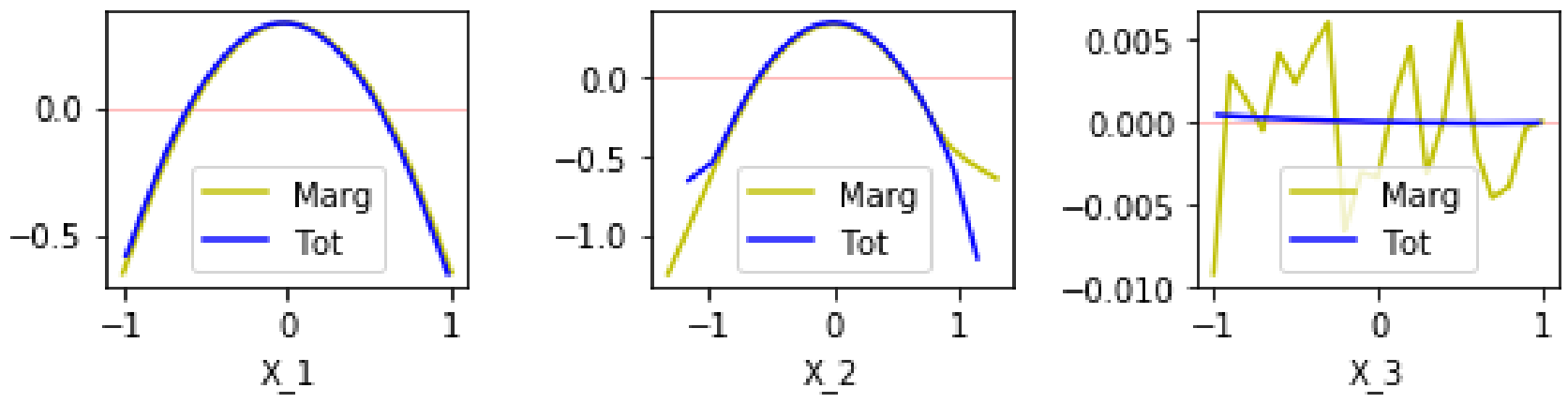

**Figure 4.b. ATDEV and marginal plot overlay**

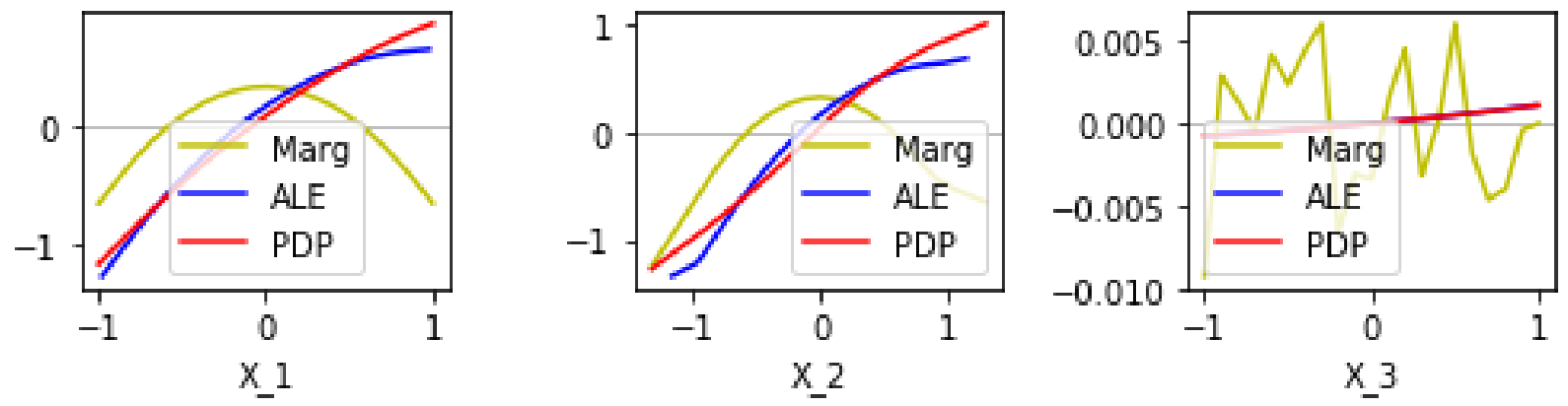

**Figure 4.c. PDPs, marginal and ALE plots overlay**

**Case 3. A more complicated example**

We generate $x_1, x_2, x_3 \sim Unif(-1,1)$, $\epsilon \sim N(0, 0.1^2)$, which are all independent; we also generate $x_2$ through the following:

$$x_4 = x_2 + \epsilon, \qquad x_5 = -x_3 + \epsilon$$

and

$$y = x_1 + \frac{1}{2}(3x_2^2 - 1) + \frac{1}{2}(4x_3^3 - 3x_3) + 0.8x_2x_4 + \epsilon$$

In this case, $x_2$ and $x_4$ have a strong positive correlation of 0.985; $x_3$ and $x_5$ have a strong negative correlation of -0.985. NN has $R^2 = 0.988$ with the validation dataset, close to the theoretical $R^2 = 0.989$.

The diagonals of ATDEV matrix plot in Figure 5.a capture the linear pattern of $x_1$, the quadratic pattern of $x_2$ and the cubic pattern of $x_3$ in the model. The quadratic individual effect of $x_4$ comes from the interaction term of $x_2x_4$. As discussed in Section 4.2, the 2-way interaction can transfer into quadratic individual effects for $x_2$ and $x_4$ due to their linear correlation. The off-diagonal subplots of (2,4) and (4,2) capture the ACEs between $x_2$ and $x_4$. And the off-diagonal of (3,5) captures ACE of $x_5$ through $x_3$ due to their negative correlation despite the fact that $x_5$ is not in the model. Note that off-diagonal of (5,3) is null because $x_5$ is not in the model. This is consistent with our explanation of off-diagonal plot shapes in Section 5.1 and the asymmetry patterns of ATDEV matrix plot in Section 5.5.

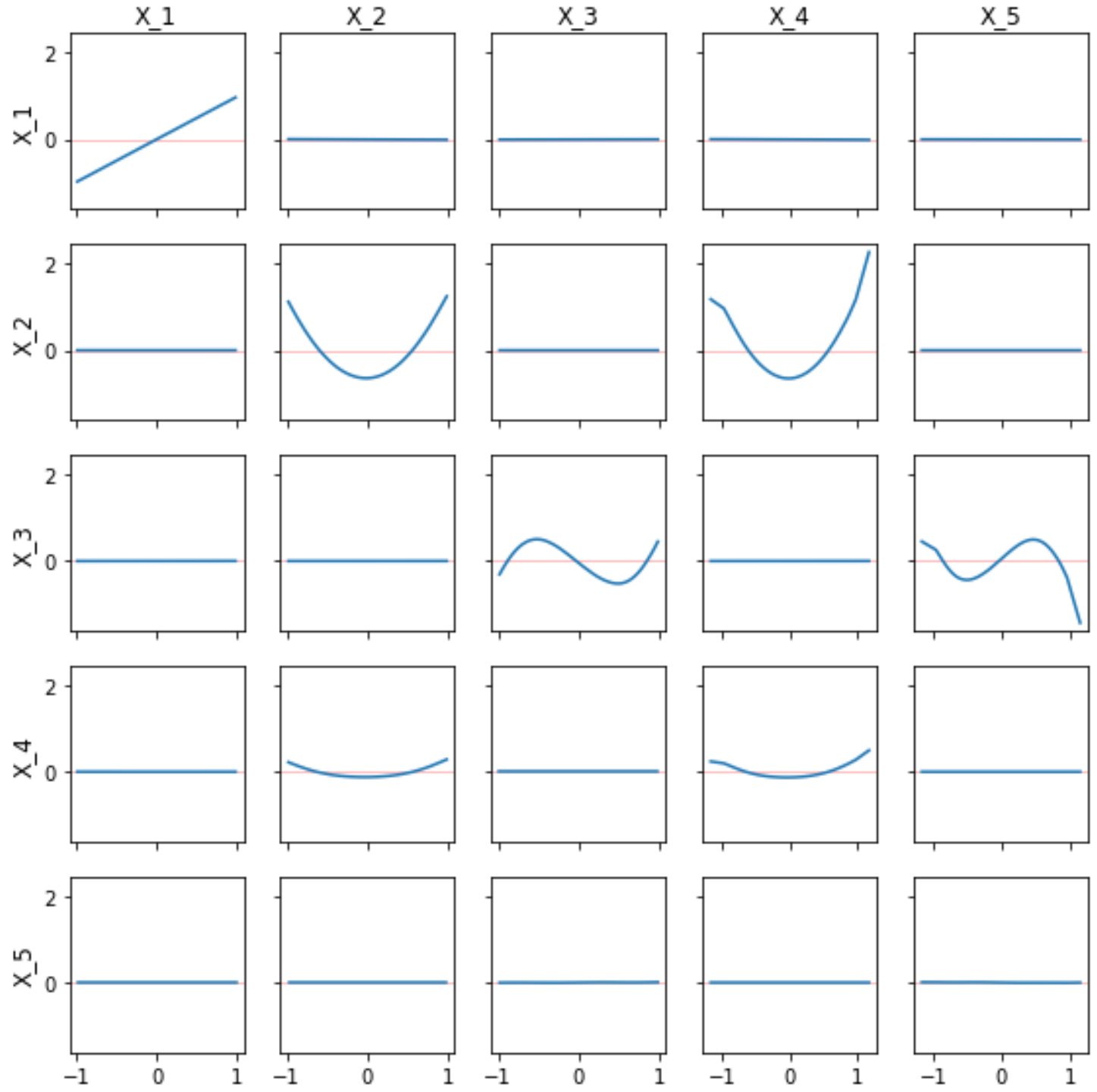

**Figure 5.a. ATDEV matrix plot**

The ATDEV and marginal plot overlay in Figure 5.b verifies that our calculation of ATDEV in this case is reliable. Note that the gaps between ATDEV plots and marginal plots for $x_4$ and $x_5$ at the right tails are caused by data sparsity at the tails due to the normally distributed random noises, which leads to numerical calculation inaccuracies for both curves.

In Figure 5.c clear extrapolations of PDP are shown for $x_4$ and $x_5$ as their shapes are inconsistent with the analytic forms, which should be 0 constant for $x_4$ and $x_5$. The inconsistency between PDP and ALE for $x_2$ may due to either PDP extrapolation or different treatments of the correlated interaction terms of PDP and ALE in their analytical forms, or both.

Figure 5.d shows an example of the comparison between ATDEV components heat map (left) and the heat map based on correlation matrix (right) as described in Section 5.4. The heat map based on correlation matrix shows the positive correlation between $x_2$ and $x_4$, and the negative correlation between $x_3$ and $x_5$. The related patterns are symmetric. However, not all the 4 off-diagonal subplots

in ATDEV components heat map are bright. Only subplots (2,4) and (3,5) are identified in the left heat map as their row variables $x_2$ and $x_3$ are significant in the model. In comparison, their mirrored subplots (4,2) and (5,3) are dark since their row variables $x_4$ and $x_5$, acting as "mediators" in ACE plots, are not that important in the model. The ATDEV components heat map is a simplified version of ATDEV matrix plot in Figure 5.a.

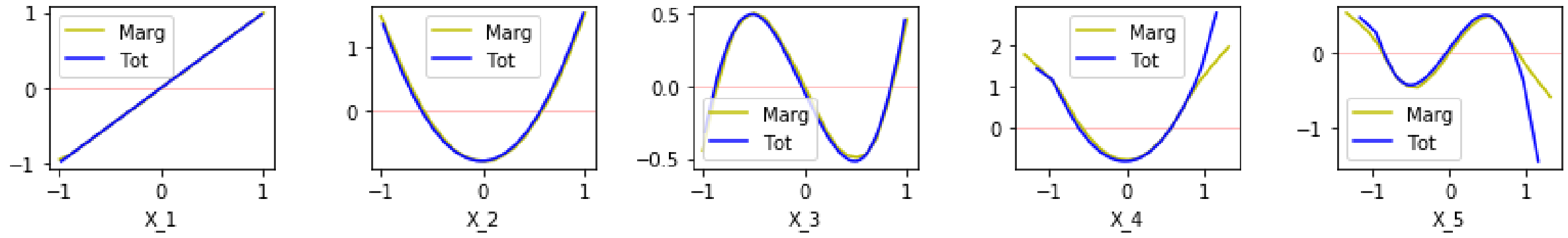

**Figure 5.b. ATDEV and marginal plot overlay**

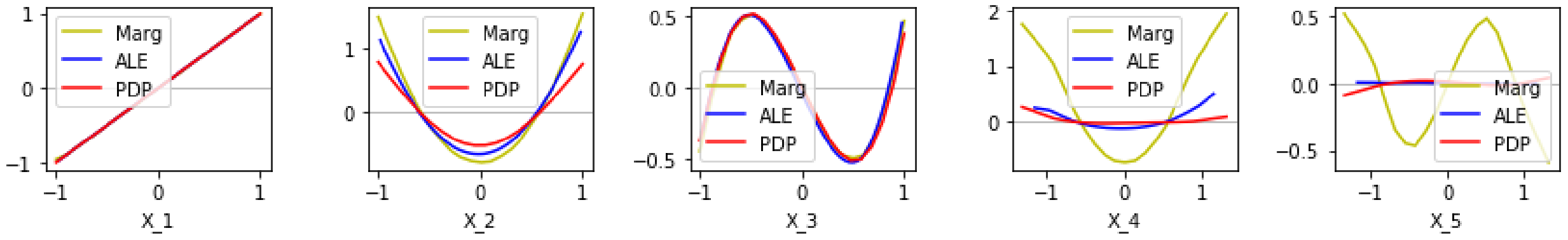

**Figure 5.c. PDPs, marginal and ALE plots overlay**

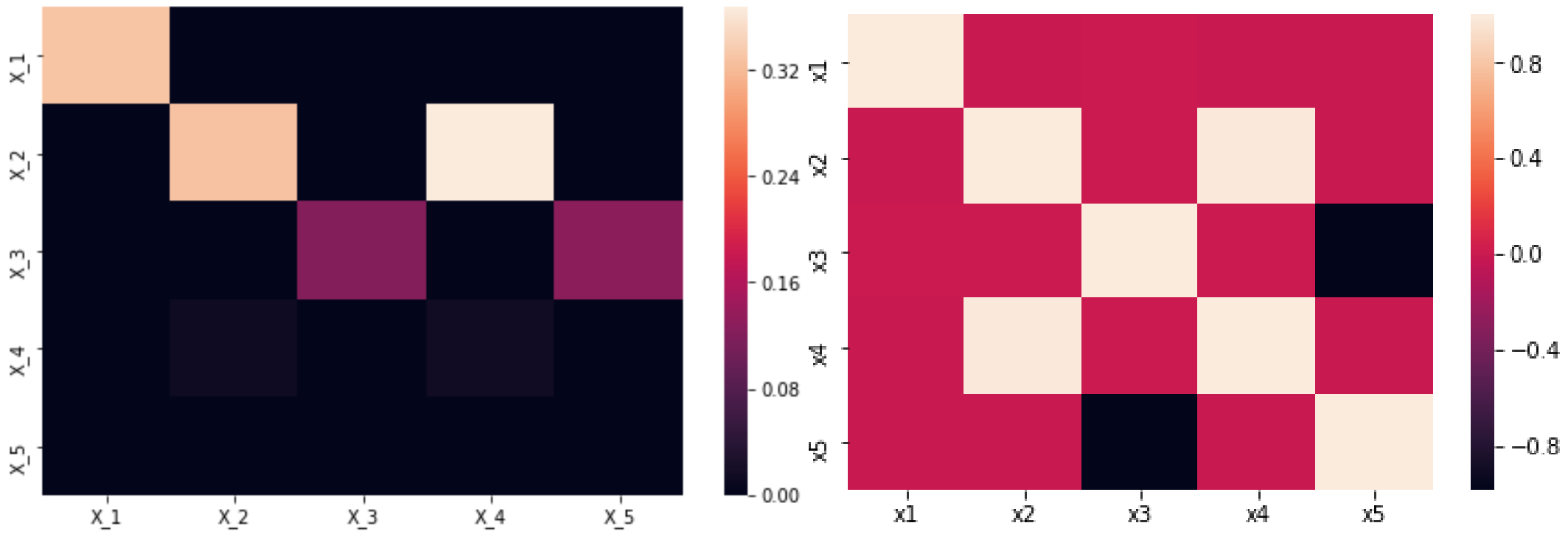

**Figure 5.d. (left) ATDEV components heat map; (right) correlation matrix heat map**

## 7. Additional Visualization Tools for Diagnostics based on Derivatives

Similar to the matrix plot of ATDEV introduced in Section 5.1, an alternative matrix plot can be created based on the conditional expectation of partial derivatives without integration (i.e., accumulation), which we simply refer to as Local Effect (LE) matrix plot. In such a matrix plot, the $j$-th diagonal plot is given by the estimate of

$$f_j^{LE}(x_j) = E_{\boldsymbol{X}_{-j}|X_j}\{f_j^1(X_j, \boldsymbol{X}_{-\boldsymbol{j}})|X_j = x_j\}, -Eq(14)$$

and the $(k, j)$th off-diagonal plot is given by the estimate of

$$f_{k,j}^{LE}(x_j) = E_{\boldsymbol{X}_{-j}|X_j}\{f_k^1(X_k, \boldsymbol{X}_{-\boldsymbol{k}})|X_j = x_j\}, \qquad i \neq j. -Eq(15)$$

Similar to $f_j^{ALE}(x_j)$, the diagonal subplots of $f_j^{LE}(x_j)$ show the individual contribution of $x_j$, but in derivative scale rather than response scale. However, LE plots are particularly useful in capturing interactions effectively when there is little dependency among the data. In addition, without integration, they are more sensitive to any small changes of the response surfaces, thus can be used as effective diagnostic tools to identify potential data or algorithm problems.

In addition to calculating LE matrix plot, the sample points of partial derivatives $f_j^1(X_j, \boldsymbol{X}_{-\boldsymbol{j}})$ can be leveraged in a number of different ways to facilitate comprehensive understanding of the black-box algorithms (Barlowe, et al., 2008). In the following paragraphs, we revisit simulation case 3 in Section 6.2 to demonstrate the use of these tools.

## 7.1. Example

We first start with the independent case by generating $x_1, x_2, x_3, x_4, x_5 \sim Unif(-1,1)$, $\epsilon \sim N(0, 0.1^2)$, which are all independent. Following the model form in Section 6.2, case 3,

$$y = x_1 + \frac{1}{2}(3x_2^2 - 1) + \frac{1}{2}(4x_3^3 - 3x_3) + 0.8x_2x_4 + \epsilon.$$

The left panel of Figure 6.a below is the LE matrix plot based on $Eq(14) - Eq(15)$. Since there is no correlation between any pair of the features, the diagonal subplots accurately capture the main effects of $x_1, x_2$ and $x_3$, and the off-diagonals also successfully capture the interaction between $x_2$ and $x_4$.

An alternative ways is to check the scatterplot of partial derivatives before taking conditional expectation, shown in the right panel of Figure 6.a blow. These plots show a similar story as suggested by LE matrix plot , but with more information from the distribution of sample points. For example, the off-diagonal subplots (2,4) and (4,2) are not symmetric. The subplot (4,2) has a strong and clear linear pattern because $f_4^1(x_4, \boldsymbol{x}_{-\boldsymbol{4}}) = 0.8x_2$, which is a linear function of $x_2$, whereas subplot (2,4) is generated from $f_2^1(x_2, \boldsymbol{x}_{-\boldsymbol{2}}) = 3x_2 + 0.8x_4$, which is a function of both $x_2$ and $x_4$, and $x_4$ plays a minor role compared with $x_2$. That is why the linear pattern in subplot (2,4) is weaker.

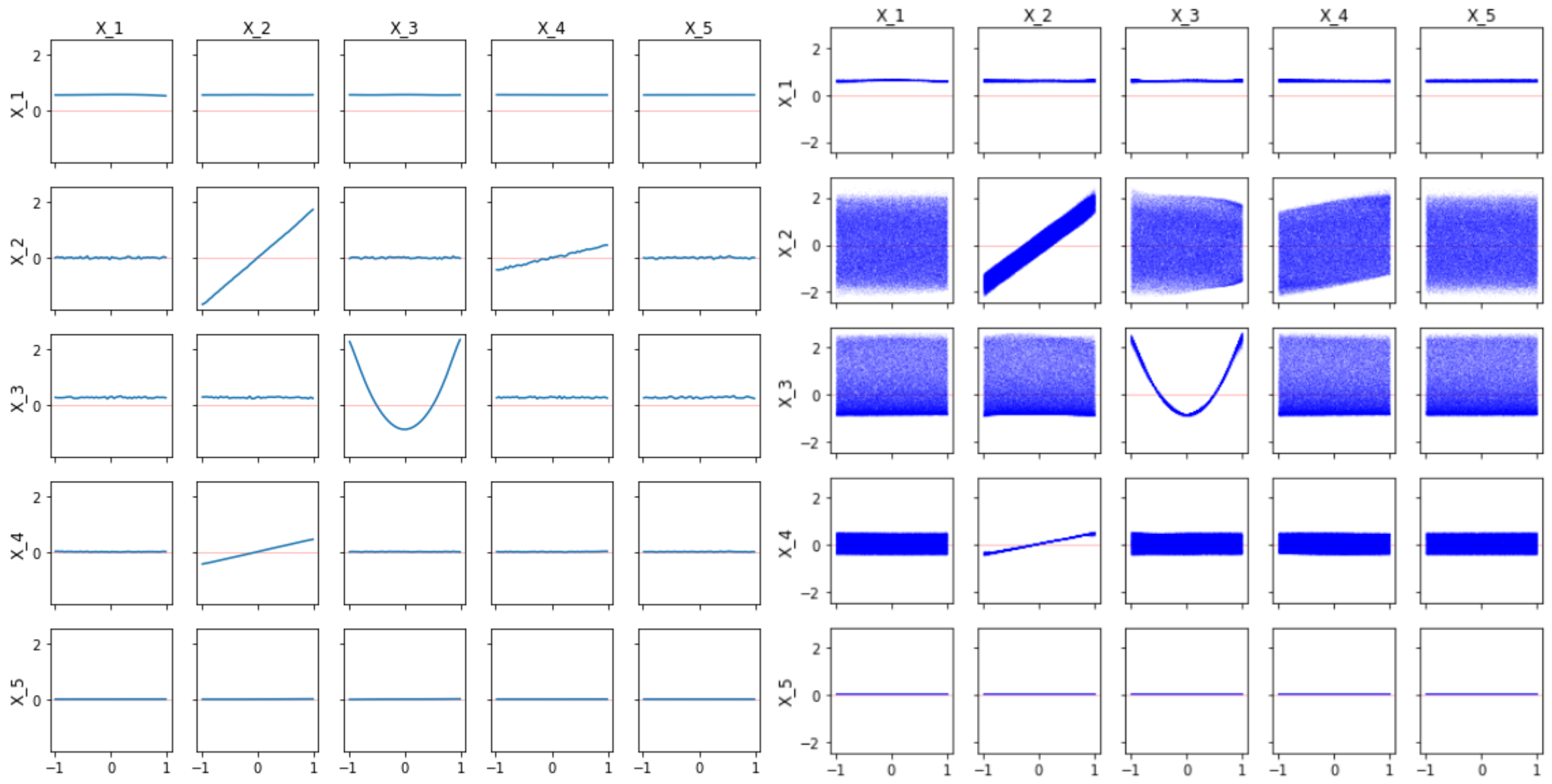

Figure 6.a. (left) LE matrix plot; (right) LE matrix scatterplot for independent data

More visualization tools can be constructed based on sample points of $f_j^1(x_j, \boldsymbol{x}_{-j})$, for example the histogram below in Figure 6.b**Error! Reference source not found.** shows the distribution of the partial derivatives.

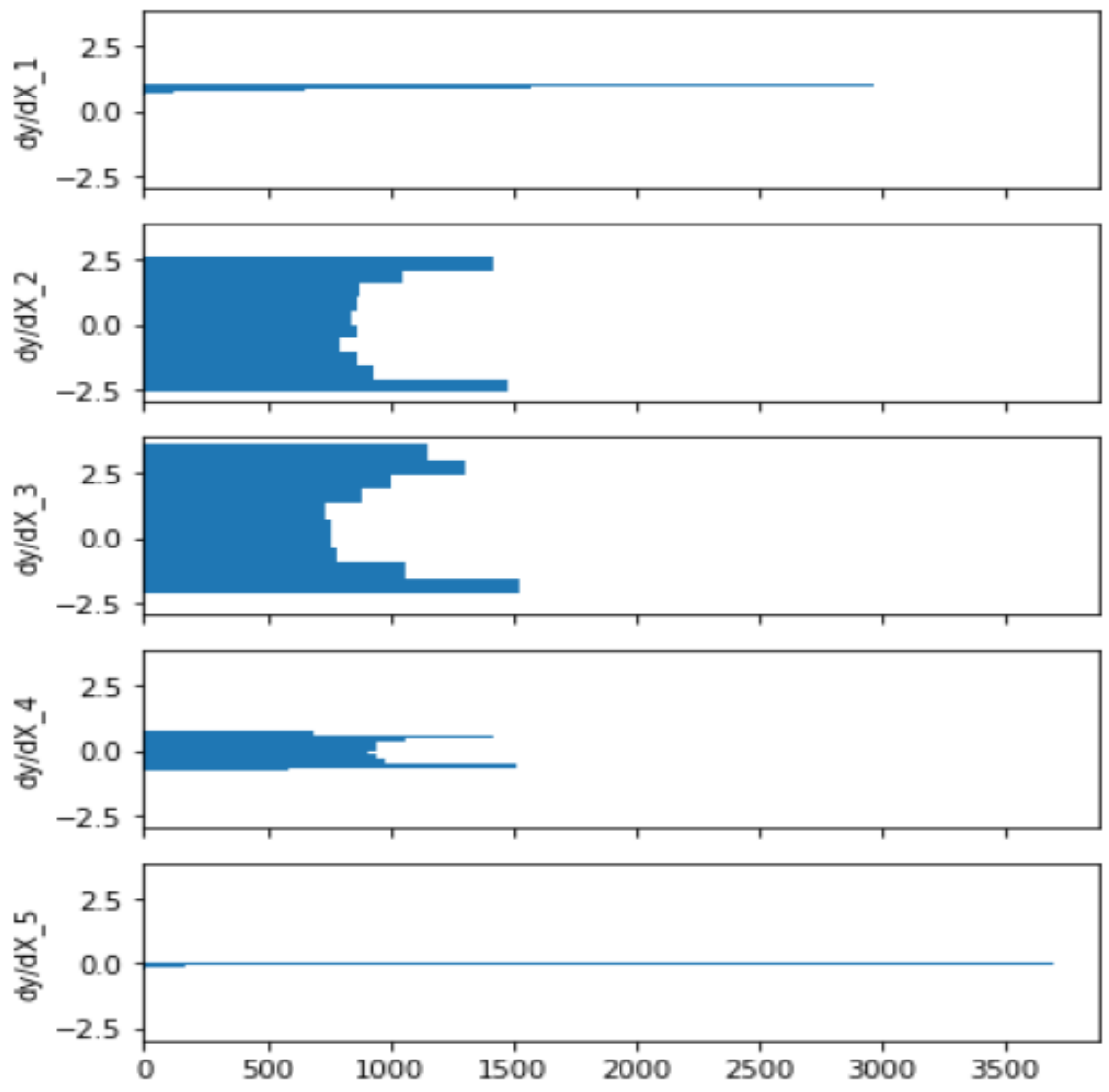

Figure 6.b. Histogram of partial derivative samples

In addition, variable importance measurement can also be constructed using partial derivatives, which is known as Derivative-based Global Sensitivity Measures (DGSM) indices in early literature (Kucherenko, 2009):

$$v_j^* = E\left\{f_j^1(X_j, \boldsymbol{X}_{-j})^2\right\}.$$

Note that $v_j^*$ is different from $v_{kj}$ and $v_{+j}$ proposed in Section 5.4, and is defined directly based on the derivatives without integration. It can be empirically estimated by taking sum of squares for each of the diagonals in the LE matrix scatterplot. In our example, the estimated DGSM from LEs can be visualized with the following bar chart in Figure 6.c, which correctly identifies $x_3$ as the most important variable in the model, followed by $x_2, x_1$ and $x_4$, and $x_5$ is not in the model.

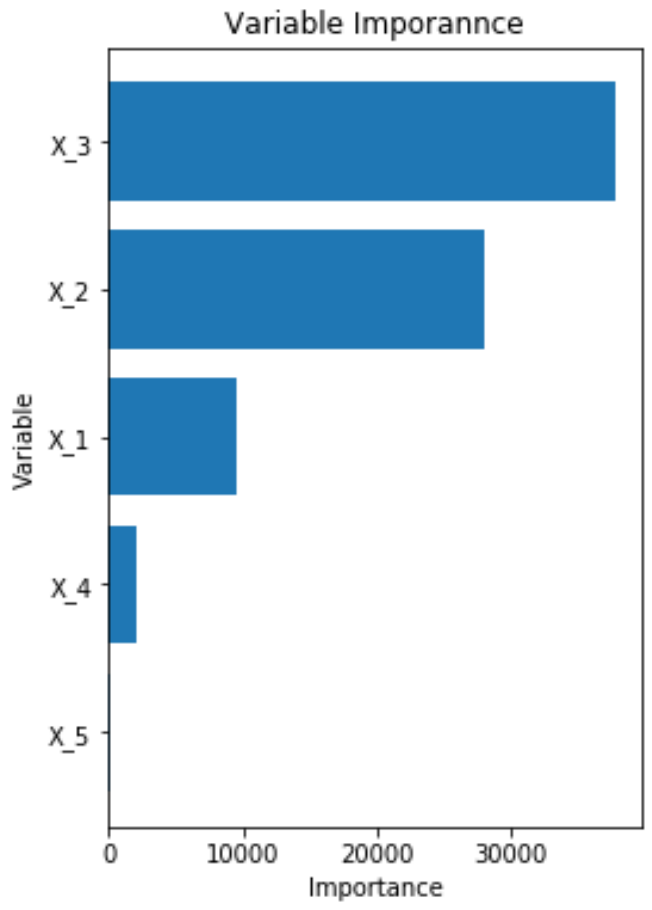

**Figure 6.c. Bar chart of DGSM for variable importance**

Note that although both the LE matrix plot and the LE matrix scatterplot are good for independent data, they can be misleading when the data are correlated, especially the off-diagonal subplots. As an illustration, we introduce strong correlations into the above simulation example between $x_2$ and $x_4$ ($\rho_{24} = 0.98$), and between $x_3$ and $x_5$ ($\rho_{35} = -0.98$). Figure 6.d below are the LE curve matrix plot and matrix scatterplot side by side.

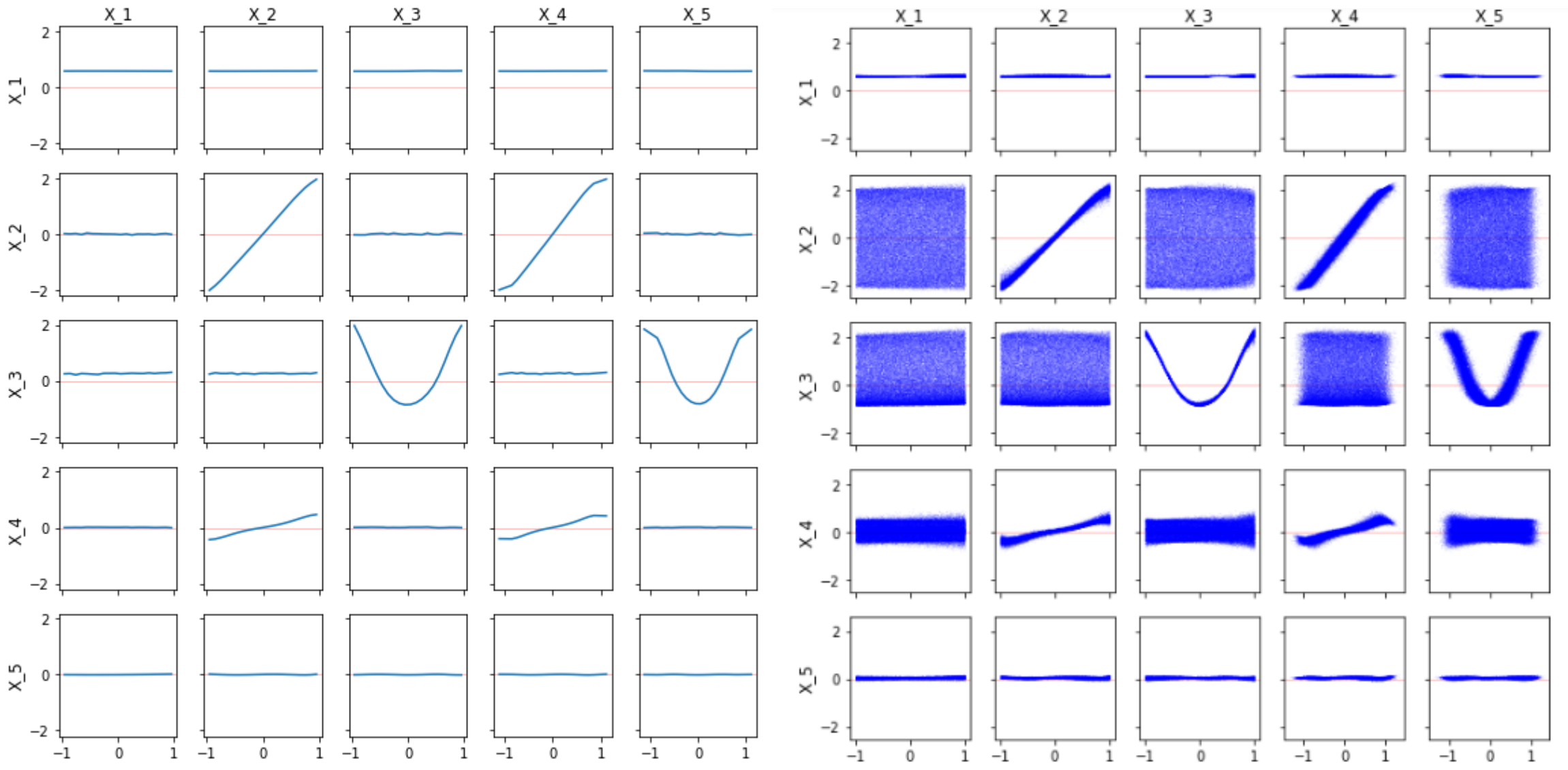

**Figure 6.d. (left) LE matrix plot; (right) LE matrix scatterplot for correlated data**

The LE matrix plots for correlated data in Figure 6.d are different from that for independent data in Figure 6.a in the following ways:

a. The diagonal plot for $x_4$ is no longer constant at 0. This is because $f_4^1(x_4, \boldsymbol{x_{-4}}) = 0.8x_2$; $x_4$ is linearly correlated with $x_2$, thus $f_4^{LE}(x_4) = 0.8E(X_2|X_4 = x_4)$ is a linear function of $x_4$.
b. The slope of off-diagonal subplot (2,4) is much steeper than the independent case, because the expectation of $f_2^1(x_2, \boldsymbol{x_{-2}}) = 3x_2 + 0.8x_4$ conditional on $x_4$ is strongly positively associated with $x_4$ due to the positive linear correlation between $x_2$ and $x_4$.
c. The off-diagonal subplot of (3,5) is no longer constant at 0, despite the fact that $x_5$ is not in the model. This is because $f_{3,5}^{LE}(x_5) = E(f_3^1(X_3, \boldsymbol{X_{-3}})|X_5 = x_5) = E\left(6X_3^2 - \frac{3}{2}\middle|X_5 = x_5\right)$ is a quadratic function $x_5$ due to the linear correlation between $x_5$ and $x_3$, $x_5$ also has an "indirect" cubic effect on the response surface though it's not in the model.

Therefore, the LE matrix plots may not be good options for correlated data because both the diagonal plots and the off-diagonal plots can be affected by correlations and can no longer capture "clean" main effect terms or interaction terms in the true model as in the independent cases. In such scenarios, we recommend to use the diagnostic tools proposed in section 5.

# 8. Discussion

## 8.1. Computation of the ATDEV plots

Computation complexity varies across the different plots. Marginal plots is the most straightforward and the fastest as it does not require any scoring of out-of-sample data or calculating derivative and integrals. The computation of PDPs, by comparison, can be quite slow due to the need to specify grid points and to score the extrapolated data points based on the grid points.

The computation of ALE, ACE and ATDEV is complicated but usually faster than PDPs. They involve several key steps:

1. Estimate the partial derivative $f_j^1(x_j, \boldsymbol{x_{-j}})$ for each in-sample data point for given $j$;
2. Calculate derivatives of $m_k^1(x_j)$ for given $k$ and $j$;
3. Estimate the conditional expectations in $Eq(7)$;
4. Integrate the averages across $x_j$.

In Step 1, we can estimate $f_j^1(x_j, \boldsymbol{x_{-j}})$ analytically or numerically. The analytical approach is restricted to the algorithms with closed-form gradients, for example, Neural Networks, where the gradient functions can be extracted from the fitted algorithms. This works only for continuous variables. For the algorithms without closed-form gradients, e.g., tree-based algorithms, one needs to approximate the derivatives using finite differences. See Apley (2016) for a detailed discussion.

It is worth mentioning that even when the prediction model itself does not have closed-form gradients (such as Gradient Boosting), one can fit a NN surrogate model to the prediction model scores, and get model performance comparable to the original prediction models. The concept of surrogate models is known as emulators in the field of computer experiments (Bastos & O'Hagan, 2009), and is referred to as model distillation (Tan, et al., 2018) or model compression (Bucilua, et al., 2006) in machine learning literature, with "born again trees" (Breiman & Shang, 1997) as one of the earliest implementations.

In Step 2, there are different ways to specify $m_k(x_j)$. In our analysis, we used a linear model for $m_k(x_j)$ and used OLS to compute the regression coefficient. In practice, one can use splines or local smoothing to fit more complicated $m_k(x_j)$'s.

The quantities in Step 3 and 4 can be calculated numerically, for instance following the description in Apley (2016).

### 8.2. Binary Predictors

When the predictors are categorical, nominal or ordered, the derivative-based approaches in this paper do not carry forward directly. The extensions will be discussed in a later paper.

## 9. Concluding Remarks

This paper uses a derivative-based approach to advance the theoretical understanding of various tools used to interpret regression models, with a particular focus on non-parametric situations including supervised machine learning algorithms. Based on these, the paper proposes a suite of visualization tools that facilitate the use of black box algorithms to better peek into the 1-D effects of the correlated variables, understand the relative importance of each decomposed effect, and also identify the possible extrapolation of PDPs.

In future work, we plan to extend the ideas to 2- and higher-dimensional spaces to obtain similar insights.

## Appendix 1. Proof of Proposition 1

Let $x_k = m_k(x_j) + e_k$ , where $e_k$ is random noise with 0 mean and independent of $x_j$. Thus

$$\begin{aligned}\boldsymbol{x}_{-j} &= (x_1, \dots, x_{j-1}, x_{j+1}, \dots, x_p) \\ &= (m_1(x_j) + e_1, \dots, m_{j-1}(x_j) + e_{j-1}, m_{j+1}(x_j) + e_{j+1}, \dots, m_p(x_j) + e_p) \\ &\triangleq \boldsymbol{m}(x_j) + \boldsymbol{e}\end{aligned}$$

For Marginal plot:

$$\begin{aligned}f_j^M(x_j) &= E\{f(X_j, \boldsymbol{X_{-j}})|X_j = x_j\} = \int_{\boldsymbol{x_{-j}}} f(x_j, \boldsymbol{x_{-j}})\, g(\boldsymbol{x_{-j}}|x_j) d\boldsymbol{x_{-j}} \\ &= \int_{\boldsymbol{e}} f(x_j, \boldsymbol{m}(x_j) + \boldsymbol{e})\, g(\boldsymbol{m}(x_j) + \boldsymbol{e}|x_j) d\boldsymbol{e} = \int_{\boldsymbol{e}} f(x_j, \boldsymbol{m}(x_j) + \boldsymbol{e})\, g(\boldsymbol{e}) d\boldsymbol{e} \\ &= \int_{\boldsymbol{e}} \left( \int_{z_{j,0}}^{x_j} \frac{df(z_j, \boldsymbol{m}(z_j) + \boldsymbol{e})}{dz_j} dz_j + C(\boldsymbol{e}) \right) g(\boldsymbol{e}) d\boldsymbol{e} \\ &= \int_{z_{j,0}}^{x_j} \int_{\boldsymbol{e}} \frac{df(z_j, \boldsymbol{m}(z_j) + \boldsymbol{e})}{dz_j} g(\boldsymbol{e})\, d\boldsymbol{e} dz_j + C = \int_{z_{j,0}}^{x_j} E_e \left\{ \frac{df(z_j, \boldsymbol{m}(z_j) + \boldsymbol{e})}{dz_j} \right\} dz_j + C \\ &= \int_{z_{j,0}}^{x_j} E_e\{f_{j,T}^1(z_j, \boldsymbol{m}(z_j) + \boldsymbol{e})\}\, dz_j + C\end{aligned}$$

Similarly, for ATDEV plot:

$$\begin{aligned}f_j^{Tot}(x_j) &= \int_{z_{j,0}}^{x_1} E\{f_{j,T}^1(X_j, \boldsymbol{X_{-j}})|X_j = z_j\}\, dz_j = \int_{z_{j,0}}^{x_1} \int_{\boldsymbol{x_{-j}}} f_{j,T}^1(z_j, \boldsymbol{x_{-j}}) g(\boldsymbol{x_{-j}}|z_j)\, d\boldsymbol{x_{-j}} dz_j \\ &= \int_{z_{j,0}}^{x_1} \int_{\boldsymbol{x_{-j}}} f_{j,T}^1(z_j, \boldsymbol{x_{-j}}) g(\boldsymbol{m}(z_j) + \boldsymbol{e}|z_j)\, d\boldsymbol{e} dz_j \\ &= \int_{z_{j,0}}^{x_1} \int_{\boldsymbol{x_{-j}}} f_{j,T}^1(z_j, \boldsymbol{m}(z_j) + \boldsymbol{e}) g(\boldsymbol{e})\, d\boldsymbol{e} dz_j = \int_{z_{j,0}}^{x_1} E_{\boldsymbol{e}}\{f_{j,T}^1(z_j, \boldsymbol{m}(z_j) + \boldsymbol{e})\}\, dz_j\end{aligned}$$

Therefore,

$$f_j^M(x_j) \equiv f_j^{Tot}(x_j) + C. \square$$